\documentclass{article}
\usepackage{iclr2027_conference,times}

\usepackage{hyperref}
\usepackage{url}
\usepackage{graphicx}
\usepackage{caption}

\usepackage{newfloat}
\usepackage{float}
\usepackage{wrapfig}
\usepackage{needspace}
\usepackage{listings}
\DeclareCaptionStyle{ruled}{labelfont=normalfont,labelsep=colon,strut=off}
\floatstyle{ruled}
\newfloat{listing}{tb}{lst}{}
\floatname{listing}{Listing}

\usepackage{booktabs}

\usepackage{array}
\usepackage[table]{xcolor}
\usepackage{xspace}
\usepackage{amsmath}
\usepackage{circledsteps}
\usepackage{enumitem}
\usepackage{multirow}
\usepackage{subcaption}
\usepackage{makecell}
\usepackage{xurl}
\usepackage{tabularx}
\usepackage{colortbl}
\newcolumntype{Y}{>{\centering\arraybackslash}X}
\usepackage{amsmath,amssymb,amsfonts}
\usepackage{textcomp}
\usepackage[framemethod=TikZ]{mdframed}

\newcommand{\todoc}[2]{{\textcolor{#1}{\textbf{#2}}}}

\definecolor{lightred}{RGB}{255, 204, 204}

\renewcommand{\todoc}[2]{\relax}

\newcommand{\code}[1]{\lstinline[basicstyle=\ttfamily\small,breaklines=true]@#1@}

\newcolumntype{a}{>{\columncolor{lightgray}}c}

\providecommand{\circled}[1]{\Circled[inner xsep=2.5pt,inner ysep=2.5pt]{#1}}

\DeclareRobustCommand{\ours}{%
    \ifmmode
        \text{T\scalebox{0.8}{ENET}}%
    \else
        T\scalebox{0.8}{ENET}%
    \fi
    \xspace
}

\newcommand{\distance}{10pt}
\usepackage{adjustbox}
\usepackage{circledsteps}

\mdfdefinestyle{findingbox}{%
    linecolor=gray!70,
    linewidth=3.5pt,
    innerleftmargin=10pt,
    innerrightmargin=10pt,
    innertopmargin=8pt,
    innerbottommargin=8pt,
    backgroundcolor=gray!10,
    topline=false,
    rightline=false,
    bottomline=false,
}

\iclrfinalcopy

\title{Analyzing and Mitigating Cost-Inefficient Behaviors in Coding Agents}

\author{Yiran Hu,
  Nan Jiang$^\dagger$,
  Shanchao Liang,
  Anik Dey,
  Yi Wu,
  and Lin Tan \\
  Purdue University \\
  {\small \{hu954, liang422, dey58, wu1827, lintan\}@purdue.edu},
  {\small $^\dagger$nan.nathan.jiang@gmail.com}
}

\begin{document}

\maketitle

\vspace{-15pt}
\begin{abstract}
\vspace{-5pt}
Although effective, coding agents often incur substantial monetary costs. Their recurring \emph{cost-inefficient behaviors} remain underexplored. We conduct a systematic study of behavioral cost inefficiencies in coding agents, analyzing 1,200 trajectories from Claude Code and Mini-SWE-Agent across four configurations on SWE-bench Verified. We identify three cost-inefficient behaviors: \emph{subsumed retrieval}, \emph{similar script generation}, and \emph{test re-execution}. We then evaluate three mitigation strategies: structure-aware retrieval, agent-synthesized skills, and developer-designed skills, over 10k trajectories on held-out SWE-bench Verified and Pro tasks.
Our main findings are:
(1) The three behaviors affect 79.00--98.00\% of coding tasks and account for up to 22.75\% of task cost. (2) Structure-aware retrieval can introduce retrieval overhead and alter agent delegation, causing inconsistent improvements in retrieval efficiency and cost increases of up to 28.14\%. (3) Agent-synthesized skills tend to produce low-level, trace-specific guidance, limiting their effectiveness and generality. (4) In contrast, developer-designed skills provide high-level, trace-agnostic guidance, reducing cost by up to 41.73\%, roughly twice the maximum gain from agent-synthesized skills.
\end{abstract}

\vspace{-14pt}
\section{Introduction}
\label{sec:intro}
\vspace{-6pt}

Coding agents have become increasingly capable, with systems such as Claude Code, Codex, and Cursor combining LLMs with tools and reusable skills~\citep{claude-code, openai-codex, cursor}. Yet, this capability comes at a steep cost. Token spending can reach millions of dollars annually~\citep{how-ai-spend-money}, and Uber reportedly exhausted its annual AI coding budget within four months~\citep{uber-ai-budget}. The pressure also affects model providers: subscribers can consume tens of thousands of dollars of model usage on a monthly plan, leaving OpenAI losing money on subscriptions and Anthropic rate-limiting Claude Code~\citep{chatgpt-pro-losses,claude-weekly-limits}. In addition, higher spending does not reliably improve performance~\citep{aiagentsmatter}. These trends make the cost efficiency of coding agents a growing concern for both individual developers and enterprises.

\begin{figure*}[t]
    \centering
    \includegraphics[width=0.85\textwidth]{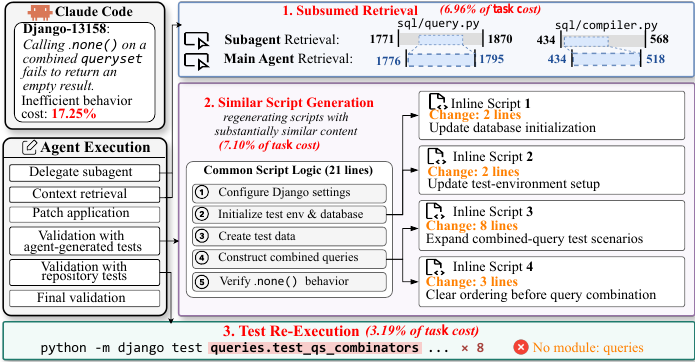}
    \captionsetup{skip=2pt}
    \caption{Cost-inefficient behaviors of Claude Code on SWE-bench Verified task \code{Django-13158}.}
    \vspace{-6pt}
    \label{fig:motivation}
\end{figure*}

Despite the effectiveness of existing cost-reduction approaches~\citep{SWE-Pruner, Toolorchestra, Graphectory}, recurring cost-inefficient behaviors during agent execution remain largely undetected, accumulating substantial end-to-end cost. Figure~\ref{fig:motivation} illustrates this problem using Claude Code on SWE-bench Verified task \code{Django-13158}~\citep{swebench-paper,swebench-verified}, which requires fixing \code{QuerySet.none()} to return an empty result on combined querysets. Claude Code delegates code exploration to a subagent, retrieves additional context, applies a patch, and validates it with generated and repository tests. Although the task passes, three cost-inefficient behaviors account for 17.25\% of its cost\footnote{Behavior cost is computed from the input, output, and cache tokens consumed by each flagged action using the corresponding model token prices (Appendix~\ref{app:behavior-cost}).}.
First, the main agent retrieves 20 lines from \code{query.py} and 85 lines from \code{compiler.py} already covered by subagent retrievals. This \textbf{\textit{subsumed retrieval}} accounts for 6.96\% of task cost. Second, it generates four similar inline test scripts sharing 21 lines of logic with minor variations. This \textbf{\textit{similar script generation}} consumes another 7.10\%. Third, it incorrectly executes \code{test_qs_combinators} eight times in the same way without updating the patch, repeatedly hitting the same module error. This \textbf{\textit{test re-execution}} costs 3.19\%. These observations motivate our first research question (\textbf{RQ1}): \textit{What cost-inefficient behavior patterns occur in coding agents?}

Among the detected behaviors, subsumed retrieval is the most prevalent. A promising remedy is structure-aware retrieval~\citep{RepoGraph, CodexGraph, LocAgent}, which enables direct access to complete code entities and explicit dependencies, reducing broad-to-narrow re-reads that cause subsumed retrieval. Although prior work reports improved localization accuracy and efficiency, whether these benefits generalize across agent architectures and translate into end-to-end cost savings remains unclear. We therefore ask \textbf{RQ2}: \textit{How does structure-aware retrieval affect cost-inefficient behaviors and the cost efficiency of coding agents?}

\vspace{-1pt}
While structure-aware retrieval primarily targets retrieval-related inefficiencies, agent skills~\citep{anthropic-skills} offer a lightweight intervention for addressing broader cost-inefficient behaviors. Such skills provide guidance that agents can apply adaptively throughout execution. We therefore study two complementary approaches: agent-synthesized skills and developer-designed skills. We ask \textbf{RQ3}: \textit{How do \textbf{agent-synthesized skills} (SynSkills) and \textbf{developer-designed skills} (DevSkills) affect cost-inefficient behaviors and the cost efficiency of coding agents?}

\vspace{-2pt}
To answer these RQs, we study cost-inefficient behaviors and their mitigation across four configurations: Claude Code (\textit{CC}) with Sonnet 4.6 (\textit{S46})~\citep{claude-sonnet-46}, and Mini-SWE-Agent (\textit{MSA})~\citep{swe-agent} with \textit{S46}, MiniMax-M3 (\textit{MM3})~\citep{minimax-m3}, or Qwen-3.5 Plus (\textit{Q35+})~\citep{qwen-35-plus}. For \textbf{RQ1}, we analyze 1,200 trajectories on 300 SWE-bench Verified tasks. For \textbf{RQ2--RQ3}, we evaluate CodeGraph~\citep{codegraph} as a representative structure-aware retrieval tool, alongside SynSkills and DevSkills, on 200 held-out Verified and 100 SWE-bench Pro tasks~\citep{swebench-pro}, with repeated runs to account for execution stochasticity.

\vspace{-2pt}
Our paper makes the following contributions:
\vspace{-5pt}
\begin{itemize}[left=0pt, itemsep=3pt, topsep=2pt, parsep=0pt, partopsep=0pt]
    \item We conduct a systematic study of behavioral cost inefficiencies in coding agents, analyzing 1,200 trajectories from four agent configurations on SWE-bench Verified. Our findings include: 
    \begin{itemize}[left=0pt, itemsep=1pt, topsep=2pt, parsep=0pt, partopsep=0pt]
        \item We identify three recurring cost-inefficient behaviors: \textbf{\textit{subsumed retrieval (SubRetrv)}}, retrieving code fully covered by prior retrievals; \textbf{\textit{similar script generation (SimScrpt)}}, regenerating highly similar scripts with only minor changes; and \textbf{\textit{test re-execution (ReTest)}}, rerunning identical tests without an updated patch. Together, they cover 79.00--98.00\% of tasks and account for 6.86--22.75\% of task monetary cost.
        \item \textbf{\textit{SubRetrv}} is the most prevalent behavior, affecting up to 92.33\% of tasks and 11.41\% of cost. SubRetrv in Claude Code mainly stems from re-reading subagent-retrieved code; in Mini-SWE-Agent, it arises from low-feedback editing, zoom-in localization, and long debugging loops.
        \item \textbf{\textit{SimScrpt}} affects up to 68.00\% of tasks and occurs 5.91--9.98$\times$ more often in Mini-SWE-Agent than in Claude Code. Built-in guidance in Claude Code largely confines SimScrpt to ephemeral scripts, while Mini-SWE-Agent also regenerates similar files for testing and editing.
        \item \textbf{\textit{ReTest}} affects 49.67--83.00\% of tasks and contributes up to 5.39\% of task cost. It primarily arises from repository-specific test knowledge gaps, test-signal recovery, and progress stalls.
    \end{itemize}

    \item We address a key gap in the evaluation of structure-aware retrieval by examining whether it consistently improves retrieval efficiency and whether such gains translate into end-to-end cost savings.
    \begin{itemize}[left=0pt, itemsep=1pt, topsep=2pt, parsep=0pt, partopsep=0pt]
        \item \textbf{Structure-aware retrieval is not inherently cost-efficient.} Depending on agent architecture and model, its verbose feedback and altered agent delegation can introduce retrieval overhead, yielding inconsistent SubRetrv reductions and end-to-end cost increases of up to 28.14\%.
    \end{itemize}

    \item We design and evaluate a lightweight agent-driven framework that synthesizes behavioral optimization skills from diagnosed cost-inefficient behaviors.
    \begin{itemize}[left=0pt, itemsep=1pt, topsep=2pt, parsep=0pt, partopsep=0pt]
    \item \textbf{Agent-synthesized skills tend to produce low-level, trace-specific guidance, limiting their effectiveness and generality.} They yield no robust cost increase and reduce cost by at most 22.32\%, roughly half the maximum achieved by developer-designed skills.
    \end{itemize}

    \item We develop a compact set of behavior-targeted optimization skills. \textbf{Our high-level, trace-agnostic developer-designed skills achieve the largest cost savings}, reducing task cost by 7.88--41.73\%, with maximum reductions of 38.57--88.91\% across the three behaviors. These results highlight the value of human insights in agent behavioral optimization.

\end{itemize}
\vspace{-2pt}
Our implementation and analysis are available in this \href{https://anonymous.4open.science/r/coding-agent-cost-study-4B4A}{anonymous repository}.

\vspace{-10pt}
\section{Study Setup}
\label{sec:study-setup}
\vspace{-10pt}
\textbf{Research Questions}.
For \textbf{RQ1}, we analyze 1,200 trajectories across four agent configurations and 300 tasks. We normalize trajectories and label actions using an author-defined taxonomy with rule-based and LLM-assisted labeling (Appendix~\ref{app:labeling}). We then define and manually calibrate detectors to identify three cost-inefficient behaviors and examine their underlying mechanisms.
For \textbf{RQ2--RQ3}, we evaluate three approaches for mitigating the cost-inefficient behaviors, comparing each with its baseline on Pass@1, task monetary cost (cost), cost-of-pass (CoP)~\citep{CoP}, and behavior changes. 
In \textbf{RQ2}, we equip agents with CodeGraph for structure-aware retrieval. \textit{CC} accesses this through an MCP tool, while \textit{MSA} uses command-line wrappers. 
For SynSkills in \textbf{RQ3}, we build a lightweight analysis agent using each configuration's backbone model to distill corrective rules from its RQ1 trajectories and apply Trace2Skill~\citep{Trace2Skill} to consolidate them into a configuration-specific skill set.
For DevSkills, we manually derive a shared skill set from the RQ1 findings under two principles: (1) generalize across configurations and (2) mitigate cost-inefficient behaviors without sacrificing task performance. Following Trace2Skill, all skill sets are preloaded into the agent system prompt for evaluation (Appendix~\ref{app:rq3}).

\textbf{Agent Configuration.}
To disentangle the effects of agent framework and backbone model on cost-inefficient behaviors, we study two agent frameworks and vary the model within one of them. \textit{CC} is a proprietary, production-grade framework with structured \texttt{Read}/\texttt{Edit} tools and subagent delegation, whereas \textit{MSA} is an open-source, single-agent framework that interacts with repositories through shell commands. We evaluate four agent configurations: \textit{CC} with \textit{S46} as the main model and Haiku 4.5 (\textit{H45}) for subagents, and \textit{MSA} with \textit{S46}, \textit{MM3}, or \textit{Q35+}. Comparing \textit{CC} with \textit{MSA\textsubscript{S46}} captures framework-associated differences under the same main model; comparing the three \textit{MSA} configurations reveals model-associated differences under a fixed framework.

\textbf{Benchmarks \& Task Split.}
To separate behavior discovery from mitigation evaluation, we split the 500 \emph{SWE-bench Verified} tasks within each repository by creation date: the earliest 300 tasks form the \textbf{RQ1} analysis set and the remaining 200 form the held-out set \emph{Verified-200} for \textbf{RQ2--RQ3}, while preserving repository distribution. 
For cross-benchmark evaluation, we sample 100 tasks from \emph{SWE-bench Pro} (\emph{Pro-100}), covering all 11 repositories and favoring costlier tasks (Appendix~\ref{app:data-select}). Together, \emph{Verified-200} and \emph{Pro-100} form the 300-task mitigation evaluation set for \textbf{RQ2--RQ3}.

\textbf{Controlling Execution Randomness.}
To account for run-to-run noise in \textbf{RQ2--RQ3}, we run eight baselines (four configurations $\times$ two benchmarks) three times and estimate noise floors using the standard deviation of Pass@1 and coefficients of variation of cost and CoP. For unreplicated approaches, we transfer the matching baseline's noise floor, following the standard equal-variance assumption that our mitigations do not alter the sampling and infrastructure nondeterminism driving run-to-run noise~\citep{llm-nondeterminism}. We validate this assumption on nine replicated approach cells spanning approaches, agent frameworks, and benchmarks; eight show no distinguishable noise amplification, while we use the larger observed noise for the exception (Appendix~\ref{app:randomness}). We classify an effect as \emph{robust} at one-sided $p<0.05$, imposing a stricter threshold on single-run effects. Because behavior-level metrics are noisier, we report their changes only as mechanistic evidence.

\vspace{-10pt}
\section{RQ1: Analyzing Cost-Inefficient Behaviors in Coding Agents} 
\label{sec:RQ1}
\vspace{-5pt}

\begin{table*}[t]
\centering
\scriptsize
\captionsetup{skip=2pt}
\caption{Prevalence, frequency, and cost contribution of cost-inefficient behaviors across agent configurations. \textbf{Tasks}: percentage of tasks exhibiting the behavior; \textbf{Freq.}: average number of behaviors per task; \textbf{Cost}: average share of per-task monetary cost attributable to the behavior.}
\label{tab:inefficiency-summary}
\setlength{\tabcolsep}{2pt}
\renewcommand{\arraystretch}{0.75}
\resizebox{0.9\textwidth}{!}{%
\begin{tabular}{@{}l | rrr | rrr | rrr |rrr@{}}
\toprule
& \multicolumn{3}{c}{\textbf{CC}}
& \multicolumn{3}{c}{\textbf{MSA\textsubscript{S46}}}
& \multicolumn{3}{c}{\textbf{MSA\textsubscript{MM3}}}
& \multicolumn{3}{c}{\textbf{MSA\textsubscript{Q35+}}} \\
\cmidrule(lr){2-4}\cmidrule(lr){5-7}\cmidrule(lr){8-10}\cmidrule(lr){11-13}
\textbf{Behaviors}
& \textbf{Tasks} & \textbf{Freq.} & \textbf{Cost}
& \textbf{Tasks} & \textbf{Freq.} & \textbf{Cost}
& \textbf{Tasks} & \textbf{Freq.} & \textbf{Cost}
& \textbf{Tasks} & \textbf{Freq.} & \textbf{Cost} \\
\midrule
SubRetrv
& 64.33\% & 2.15 & 5.01\%
& 87.33\% & 3.45 & 8.42\%
& 92.33\% & 5.14 & 7.88\%
& 89.33\% & 6.37 & 11.41\% \\
SimScrpt
& 20.67\% & 0.43 & 1.02\%
& 51.33\% & 2.54 & 9.57\%
& 68.00\% & 4.29 & 8.85\%
& 57.67\% & 3.20 & 7.91\% \\
ReTest
& 49.67\% & 2.09 & 0.83\%
& 69.00\% & 2.51 & 3.17\%
& 83.00\% & 5.29 & 5.39\%
& 66.00\% & 2.10 & 3.43\% \\
\midrule
\textbf{Total}
& \textbf{79.00\%} & \textbf{4.67} & \textbf{6.86\%}
& \textbf{97.33\%} & \textbf{8.50} & \textbf{21.16\%}
& \textbf{98.00\%} & \textbf{14.72} & \textbf{22.12\%}
& \textbf{96.67\%} & \textbf{11.68} & \textbf{22.75\%} \\
\bottomrule
\end{tabular}
}
\vspace{1pt}
\end{table*}

We identify \textbf{three} cost-inefficient behaviors across \textit{information retrieval, script management, and solution validation} in coding agent execution. (1) \textbf{Subsumed Retrieval:} retrieving code fully covered by previous retrievals.
(2) \textbf{Similar Script Generation:} regenerating highly similar scripts with minor changes.
(3) \textbf{Test Re-Execution:} re-executing identical tests without updated patches.

\begin{wrapfigure}{r}{0.46\textwidth}
\vspace{-4pt}
\centering
\includegraphics[width=\linewidth,
  trim={0pt 3pt 0pt 5pt},
  clip]{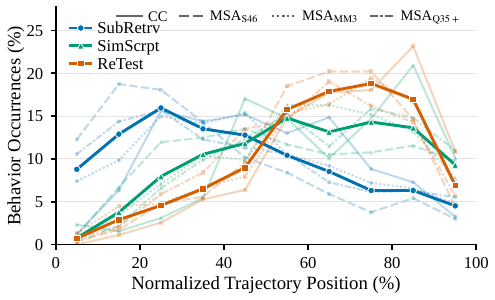}
\captionsetup{
    width=0.95\linewidth,
    justification=raggedright,
    singlelinecheck=false
}
\captionsetup{skip=4pt}
\caption{Temporal distribution of the cost-inefficient behaviors. Bold curves aggregate configurations; thin curves show individual ones.}
\label{fig:behavior-position}
\vspace{-8pt}
\end{wrapfigure}

Table~\ref{tab:inefficiency-summary} reveals that the three behaviors are both prevalent and costly, collectively affecting at least 79.00\% of tasks and accounting for up to 22.75\% of task cost, and \textit{CC} consistently exhibits the lowest overall level of inefficiency (Appendix~\ref{app:rq1-cc-better}). Figure~\ref{fig:behavior-position} further shows their temporal patterns: SubRetrv concentrates in the first half of the trajectory, coinciding with the retrieval-heavy stage; ReTest shifts toward the middle-to-late stages, where the trajectory becomes more test-heavy for patch validation; and SimScrpt is distributed more broadly rather than confined to a specific phase. We next formalize each behavior and analyze its detection, severity, and underlying mechanisms.

\vspace{-10pt}
\subsection{Subsumed Retrieval}
\vspace{-5pt}
\textbf{Detection and Severity.}
For each retrieval $b$ returning at least five non-empty lines of content, we scan preceding retrieval actions backward and identify the first action $a$ whose returned context fully covers that of $b$. We treat $(a,b)$ as a SubRetrv pair and flag $b$ as one SubRetrv action. 
Table~\ref{tab:inefficiency-summary} shows that SubRetrv is the most prevalent cost-inefficient behavior, affecting 64.33--92.33\% of tasks, occurring 2.15--6.37 times per task, and accounting for 5.01--11.41\% of task cost.

\begin{wraptable}{r}{0.51\textwidth}
\vspace{-10pt}
\centering
\scriptsize
\setlength{\tabcolsep}{2pt}
\renewcommand{\arraystretch}{0.8}
\captionsetup{
    width=0.95\linewidth,
    justification=raggedright,
    singlelinecheck=false
}
\captionsetup{skip=3pt}
\caption{SubRetrv actions by scenario, assigned in top-to-bottom priority order. Subscripts indicate the percentage of SubRetrv actions.}
\label{tab:redundant-retrieval-reasons}
\begin{tabular}{@{}l rrrr@{}}
\toprule
& \multicolumn{4}{c}{\textbf{SubRetrv Actions}} \\
\cmidrule(l){2-5}
\textbf{Occurrence Scenario} & \textbf{CC} & \textbf{MSA\textsubscript{S46}} & \textbf{MSA\textsubscript{MM3}} & \textbf{MSA\textsubscript{Q35+}} \\
\midrule
Cross-Agent SubRetrv & \textbf{324\textsubscript{50.15\%}} & — & — & — \\
Patch-Adjacent SubRetrv & 56\textsubscript{8.67\%} & 274\textsubscript{26.47\%} & 277\textsubscript{17.96\%} & 374\textsubscript{19.56\%} \\
Within-Sequence SubRetrv & 171\textsubscript{26.47\%} & \textbf{311\textsubscript{30.05\%}} & 255\textsubscript{16.54\%} & \textbf{897\textsubscript{46.91\%}} \\
Long-Distance SubRetrv & 73\textsubscript{11.30\%} & 275\textsubscript{26.57\%} & \textbf{808\textsubscript{52.40\%}} & 472\textsubscript{24.69\%} \\
Other & 22\textsubscript{3.41\%} & 175\textsubscript{16.91\%} & 202\textsubscript{13.10\%} & 169\textsubscript{8.84\%} \\
\bottomrule
\end{tabular}
\vspace{-8pt}
\end{wraptable}

\textbf{Underlying Mechanisms.} To study when and why SubRetrv occurs and how its mechanisms differ between \textit{CC} and \textit{MSA}, we classify each SubRetrv action into different scenarios in Table~\ref{tab:redundant-retrieval-reasons}. The results reveal substantial differences across agent architectures. \textit{Cross-Agent SubRetrv}, where a main-agent retrieval is subsumed by a previous subagent retrieval, is unique to \textit{CC} and accounts for 50.15\% of its SubRetrv. \textit{CC} offloads exploration to subagents, which return summaries rather than the retrieved code. So the main agent may re-retrieve the same region when needed later, as illustrated in Figure~\ref{fig:motivation}.

\phantomsection
\label{para:patch-adjacent-sr}
\textit{Patch-Adjacent SubRetrv} occurs when the SubRetrv immediately precedes or follows a patch. It is more common in \textit{MSA} (274--374 actions) than in \textit{CC} (56) due to different editing interfaces. \textit{MSA}'s shell-based editing, such as \texttt{sed -i} or generated scripts, may provide little feedback or fail silently, prompting agents to re-read code for localization or post-edit inspection. In contrast, \textit{CC}'s retrieval tools provide line numbers, while its editing tools report detailed edit outcomes, including the updated code, reducing such scenarios.

\textit{Within-Sequence SubRetrv} occurs when both retrievals in a SubRetrv pair fall within an uninterrupted retrieval sequence. It typically follows a locate-then-zoom pattern, where the agent first reads a broad region and then retrieves a subregion for precise localization. This is common in \textit{MSA}, where commands such as \code{cat} omit line numbers and often trigger follow-up \code{grep -n} or \code{sed -n} calls; \textit{CC}'s \code{Read} already returns line-numbered code, reducing this need  (171 vs.\ 255--897 actions).

\textit{Long-Distance SubRetrv} occurs when both retrievals in a SubRetrv pair are separated by at least 10 steps. Manual inspection shows that it typically reflects longer debugging loops, where earlier context becomes less salient and is re-retrieved. It is most prominent in \textit{MSA\textsubscript{MM3}} (52.40\%), which is consistent with its longest average trajectory length (77.99 steps vs.\ 29.99--58.06 for others).

\begin{mdframed}[style=findingbox]
\textbf{Finding 1.} \textit{Subsumed Retrieval} is the most prevalent behavior, affecting 64.33--92.33\% of tasks and contributing 5.01--11.41\% of task cost. Its causes are architecture-dependent: Cross-Agent SubRetrv dominates in \textit{CC}, while \textit{MSA} mainly suffers from low-feedback editing, zoom-in localization, and long debugging loops. 
\end{mdframed}

\vspace{-10pt}
\subsection{Similar Script Generation}
\vspace{-5pt}

\textbf{Behavior Detection.}
We define SimScrpt as regenerating near-duplicate scripts with minor differences rather than editing existing scripts, wasting output tokens and forgoing artifact reuse. We extract \emph{ephemeral scripts} (e.g., \code{python -c} and heredocs) and \emph{file-based scripts} (\code{.py} files) from agent actions, remove blank and comment-only lines, and retain scripts with at least five lines. For each generation action $b$, we identify the nearest preceding action $a$ whose script has line-level Jaccard similarity\footnote{We select $0.60$ by manually inspecting 20 SimScrpt pairs per configuration. Lower thresholds often match generic scaffolding, while $0.60$ retains highly similar task-specific logic.} of at least $0.60$. We treat $(a,b)$ as a SimScrpt pair and count $b$ as one occurrence.

\begin{figure}[t]
    \centering
    \begin{minipage}[t]{0.32\textwidth}
        \centering
        \includegraphics[width=\linewidth,trim={5pt 2pt 0pt 4pt}, clip]{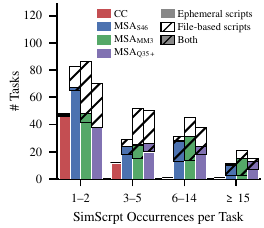}
        \captionsetup{skip=4pt}
        \caption{Distribution of SimScrpt occurrences per task.}
        \label{fig:HSSR-per-task}
    \end{minipage}
    \hfill
    \begin{minipage}[t]{0.32\textwidth}
        \centering
        \includegraphics[width=\linewidth, trim={5pt 2pt 0pt 4pt}, clip]{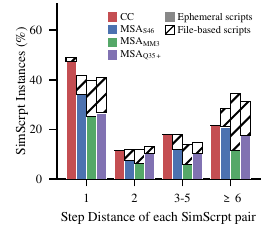}
        \captionsetup{skip=4pt}
        \caption{Distribution of SimScrpt pair step distances.}
        \label{fig:HSSR-step-distance}
    \end{minipage}
    \hfill
    \begin{minipage}[t]{0.34\textwidth}
        \centering
        \includegraphics[width=\linewidth, trim={5pt 1pt 0pt 1pt}, clip]{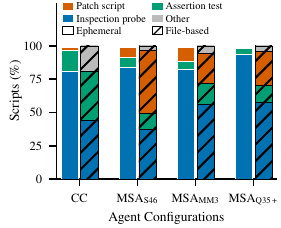}
        \captionsetup{skip=4pt}
        \caption{Functions of ephemeral and file-based SimScrpt scripts.}
        \label{fig:hssr-purpose}
    \end{minipage}
\end{figure}

\textbf{Prevalence and Severity.}
Table~\ref{tab:inefficiency-summary} shows that 
SimScrpt affects up to 68.00\% of tasks and contributes up to 9.57\% of task cost. 
Interestingly, SimScrpt is substantially more severe in \textit{MSA}, affecting 51.33--68.00\% of tasks and contributing 7.91--9.57\% of task cost, versus 20.67\% and 1.02\% in \textit{CC}. It also occurs 5.91--9.98$\times$ more frequently in \textit{MSA} (2.54--4.29 vs.\ 0.43). Figure~\ref{fig:HSSR-per-task} further reveals a pronounced high-frequency tail in \textit{MSA}: 42--66 tasks per configuration exhibit at least six SimScrpt occurrences, compared with only two in \textit{CC}.

\textbf{Underlying Mechanisms.}
\emph{When and why SimScrpt occurs.}
SimScrpt arises during issue reproduction, code editing, and patch validation, explaining its broad distribution in Figure~\ref{fig:behavior-position}. Figure~\ref{fig:HSSR-step-distance} further reveals two patterns: 39.89--48.84\% of occurrences immediately follow a similar script, typically reflecting minor revisions with full-script regeneration, while 21.71--34.60\% occur at a step distance of at least six.
The underlying causes differ by script form. Ephemeral scripts leave no reusable artifact, requiring agents to regenerate similar code when needed again. File-based scripts persist on disk, but become less salient to the agent as its trajectory grows. Reusing them requires re-reading files for precise editing, while shell-based edits such as \code{sed -i} are error-prone, as discussed in \hyperref[para:patch-adjacent-sr]{\textit{Patch-Adjacent SubRetrv}}. Regenerating a near-duplicate script can therefore become the easier immediate choice. 

\emph{Ephemeral vs.\ file-based SimScrpt scripts.}
To understand their differences, we classify SimScrpt scripts by their functionalities using abstract syntax tree (AST) analysis (Figure~\ref{fig:hssr-purpose}). A \emph{patch script} modifies source files; an \emph{assertion test} performs explicit pass/fail validation; and an \emph{inspection probe} queries or prints runtime information, such as object values, attributes, or program state. Scripts are assigned by priority: patch, test, then probe.
Ephemeral SimScrpt scripts overwhelmingly consist of inspection probes (81.03--93.45\%). Agents commonly use \code{python -c} and heredocs as disposable probes: they generate a script, inspect certain output, and discard it, then regenerate a similar probe when the information is needed again.
File-based SimScrpt scripts are more functionally diverse: inspection probes decrease, while assertion tests and, in \textit{MSA}, patch scripts account for larger shares. Because these files persist and encode reusable validation or editing logic, regenerating them represents a clearer missed opportunity for artifact reuse. 

\emph{Why SimScrpt script forms differ across architectures.}
Figures~\ref{fig:HSSR-per-task} and~\ref{fig:HSSR-step-distance} show that \textit{CC} primarily exhibits ephemeral SimScrpt, whereas \textit{MSA} exhibits both script forms. This difference is consistent with \textit{CC}'s built-in instruction: ``NEVER create files unless they're absolutely necessary for achieving your goal.'' This discourages persistent script creation and shifts toward ephemeral generation, whereas \textit{MSA} has no such constraint.
Patch scripts are rare in \textit{CC} because its built-in \code{Edit} and \code{Write} tools directly support modification action, whereas \textit{MSA}'s poor feedback of shell-based editing more often triggers agents to generate their own editing scripts.

\begin{mdframed}[style=findingbox]
\textbf{Finding 2.} \textit{Similar Script Generation} affects up to 68.00\% of tasks and contributes up to 9.57\% of task cost, occurring 5.91--9.98$\times$ more frequently in \textit{MSA} than in \textit{CC}. \textit{CC}'s built-in guidance largely confines SimScrpt to ephemeral scripts, whereas \textit{MSA} also repeatedly regenerates file-based testing and editing scripts. These patterns suggest that an ideal agent should persist reusable logic and use inline scripts for simple one-off probes.
\end{mdframed}

\vspace{-10pt}
\subsection{Test Re-Execution}
\vspace{-5pt}
\textbf{Detection and Severity.}
We partition trajectories into inter-patch windows because a new patch may justify rerunning tests, whereas repeated executions without a patch update validate the same code state. Within each window, we group executions by the executed tests, retain the final execution as potentially decision-relevant, and flag preceding ones as ReTest. 
\emph{Table~\ref{tab:inefficiency-summary} shows that ReTest affects 49.67--83.00\% of tasks and contributes up to 5.39\% of task cost}. \textit{MSA\textsubscript{MM3}} has the highest prevalence and frequency, averaging 5.29 occurrences per task, over twice that of any other configuration.

\textbf{Underlying Mechanisms.}
We identify three common causes of ReTest. (1) \emph{Repository-specific test knowledge gaps} arise when agents are unfamiliar with the proper test runner or configuration, causing repeated failures. (2) \emph{Test-signal recovery} occurs when test output is truncated, ambiguous, or poorly captured by the agent, prompting reruns to recover debugging information. (3) \emph{Progress stalls} arise when agents enter long reasoning loops without updating the patch, repeatedly reconsidering the same diagnosis and rerunning tests despite sufficient prior signals. Such behavior reflects stalled decision-making without advancing the solution.

\begin{mdframed}[style=findingbox, skipbelow=-5pt]
\textbf{Finding 3.} \textit{Test Re-Execution} affects 49.67--83.00\% of tasks and accounts for up to 5.39\% of task cost, with \textit{MSA\textsubscript{MM3}} exhibiting the highest prevalence and frequency. ReTest commonly arises from repository-specific test knowledge gaps, test-signal recovery, and progress stalls.
\end{mdframed}
\vspace{-10pt}
More detection details, analysis, and examples for RQ1 are provided in Appendix~\ref{app:rq1}.

\providecommand{\pct}[1]{\textsubscript{#1}}
\definecolor{cDown}{HTML}{0072B2}
\definecolor{cUp}{HTML}{D2506A}
\definecolor{cNoise}{HTML}{FFFFFF}

\begin{table*}[t]
\captionsetup{skip=2pt}
\caption{Effects of the three approaches on Pass@1, average cost, CoP, and the occurrence of identified behaviors. Subscripts show changes from baseline. For Pass@1, cost, and CoP, {\setlength{\fboxsep}{0pt}\colorbox{cDown!45}{\strut\,blue\,}}/{\setlength{\fboxsep}{0pt}\colorbox{cUp!45}{\strut\,red\,}} indicate robust improvements/regressions, while uncolored cells indicate changes within noise; for behaviors, blue/red indicate fewer/more instances.}
\label{tab:rq234-main-general}
\centering
\scriptsize
\setlength{\tabcolsep}{2pt}
\renewcommand{\arraystretch}{0.9}
\begin{adjustbox}{max width=1\textwidth}
\begin{tabular}{ll*{8}{r}}
\toprule
& & \multicolumn{4}{c}{Verified-200} & \multicolumn{4}{c}{Pro-100} \\[-2pt]
\cmidrule(lr){3-6}\cmidrule(lr){7-10}
Approach & Metric & \multicolumn{1}{c}{CC} & \multicolumn{1}{c}{MSA\textsubscript{S46}} & \multicolumn{1}{c}{MSA\textsubscript{MM3}} & \multicolumn{1}{c}{MSA\textsubscript{Q35+}} & \multicolumn{1}{c}{CC} & \multicolumn{1}{c}{MSA\textsubscript{S46}} & \multicolumn{1}{c}{MSA\textsubscript{MM3}} & \multicolumn{1}{c}{MSA\textsubscript{Q35+}} \\[-2pt]
\midrule
\multirow{6}{*}{CodeGraph}
 & Pass@1 (pp) & \cellcolor{cNoise}74.17\pct{-0.83} & \cellcolor{cNoise}72.50\pct{+0.00} & \cellcolor{cNoise}75.50\pct{+0.33} & \cellcolor{cDown!18}72.50\pct{+5.33} & \cellcolor{cNoise}55.17\pct{+1.92} & \cellcolor{cNoise}51.72\pct{+1.92} & \cellcolor{cNoise}57.47\pct{-1.15} & \cellcolor{cNoise}39.08\pct{-3.45} \\
 & Cost (\$) & \cellcolor{cNoise}0.596\pct{+8.30\%} & \cellcolor{cNoise}0.619\pct{-3.03\%} & \cellcolor{cUp!36}0.591\pct{+28.14\%} & \cellcolor{cNoise}0.112\pct{-3.29\%} & \cellcolor{cUp!27}1.061\pct{+12.19\%} & \cellcolor{cNoise}1.132\pct{-0.25\%} & \cellcolor{cUp!18}0.985\pct{+8.39\%} & \cellcolor{cUp!27}0.150\pct{+12.99\%} \\
 & CoP & \cellcolor{cNoise}0.804\pct{+10.22\%} & \cellcolor{cNoise}0.854\pct{-3.04\%} & \cellcolor{cUp!36}0.782\pct{+27.52\%} & \cellcolor{cNoise}0.155\pct{-10.63\%} & \cellcolor{cNoise}1.923\pct{+8.69\%} & \cellcolor{cNoise}2.188\pct{-4.32\%} & \cellcolor{cNoise}1.715\pct{+10.20\%} & \cellcolor{cUp!36}0.388\pct{+23.20\%} \\
 \cmidrule(lr){2-10}
 & SubRetrv & \cellcolor{cDown!70}0.16\pct{-84.18\%} & \cellcolor{cDown!36}2.87\pct{-25.33\%} & \cellcolor{cUp!27}6.36\pct{+18.01\%} & \cellcolor{cDown!36}4.65\pct{-24.25\%} & \cellcolor{cDown!70}0.38\pct{-75.72\%} & \cellcolor{cUp!10}2.90\pct{+3.33\%} & \cellcolor{cUp!46}6.34\pct{+34.74\%} & \cellcolor{cUp!27}4.51\pct{+13.01\%} \\
 & SimScrpt & \cellcolor{cDown!36}0.26\pct{-25.85\%} & \cellcolor{cDown!27}2.90\pct{-17.01\%} & \cellcolor{cUp!36}5.90\pct{+28.49\%} & \cellcolor{cDown!46}2.68\pct{-39.61\%} & \cellcolor{cDown!10}0.10\pct{-0.21\%} & \cellcolor{cUp!70}1.64\pct{+63.26\%} & \cellcolor{cDown!10}1.99\pct{-1.33\%} & \cellcolor{cDown!57}0.79\pct{-53.46\%} \\
 & ReTest & \cellcolor{cUp!27}1.75\pct{+13.85\%} & \cellcolor{cDown!18}2.05\pct{-8.07\%} & \cellcolor{cUp!18}5.55\pct{+7.34\%} & \cellcolor{cUp!10}2.21\pct{+0.30\%} & \cellcolor{cDown!10}0.56\pct{-0.89\%} & \cellcolor{cUp!27}2.13\pct{+11.27\%} & \cellcolor{cUp!36}4.49\pct{+25.45\%} & \cellcolor{cUp!18}0.78\pct{+6.90\%} \\
\midrule
\multirow{6}{*}{SynSkills}
 & Pass@1 (pp) & \cellcolor{cNoise}74.50\pct{-0.50} & \cellcolor{cNoise}72.00\pct{-0.50} & \cellcolor{cNoise}75.50\pct{+0.33} & \cellcolor{cNoise}69.67\pct{+2.50} & \cellcolor{cNoise}55.00\pct{+0.67} & \cellcolor{cNoise}52.00\pct{+0.00} & \cellcolor{cNoise}62.00\pct{+2.00} & \cellcolor{cNoise}46.00\pct{+1.00} \\
 & Cost (\$) & \cellcolor{cNoise}0.541\pct{-1.38\%} & \cellcolor{cDown!36}0.496\pct{-22.32\%} & \cellcolor{cNoise}0.462\pct{+0.15\%} & \cellcolor{cNoise}0.115\pct{-0.70\%} & \cellcolor{cDown!18}0.826\pct{-8.86\%} & \cellcolor{cNoise}1.006\pct{-6.13\%} & \cellcolor{cDown!27}0.766\pct{-11.90\%} & \cellcolor{cNoise}0.132\pct{-2.22\%} \\
 & CoP & \cellcolor{cNoise}0.726\pct{-0.12\%} & \cellcolor{cDown!36}0.689\pct{-21.79\%} & \cellcolor{cNoise}0.611\pct{-0.33\%} & \cellcolor{cNoise}0.166\pct{-4.40\%} & \cellcolor{cDown!18}1.503\pct{-9.48\%} & \cellcolor{cNoise}1.936\pct{-6.40\%} & \cellcolor{cNoise}1.235\pct{-15.09\%} & \cellcolor{cNoise}0.286\pct{-4.51\%} \\
 \cmidrule(lr){2-10}
 & SubRetrv & \cellcolor{cDown!46}0.56\pct{-44.62\%} & \cellcolor{cDown!27}3.40\pct{-11.38\%} & \cellcolor{cDown!36}4.17\pct{-22.56\%} & \cellcolor{cDown!18}5.83\pct{-5.05\%} & \cellcolor{cDown!46}0.98\pct{-34.31\%} & \cellcolor{cUp!10}2.67\pct{+2.50\%} & \cellcolor{cDown!27}3.76\pct{-14.29\%} & \cellcolor{cUp!10}3.86\pct{+3.21\%} \\
 & SimScrpt & \cellcolor{cDown!46}0.20\pct{-41.85\%} & \cellcolor{cDown!57}1.92\pct{-45.10\%} & \cellcolor{cDown!27}3.98\pct{-13.43\%} & \cellcolor{cDown!27}3.66\pct{-17.63\%} & \cellcolor{cDown!27}0.08\pct{-11.28\%} & \cellcolor{cUp!70}1.60\pct{+79.32\%} & \cellcolor{cDown!46}1.19\pct{-35.44\%} & \cellcolor{cDown!18}1.41\pct{-5.87\%} \\
 & ReTest & \cellcolor{cUp!27}1.75\pct{+13.20\%} & \cellcolor{cDown!46}1.46\pct{-34.53\%} & \cellcolor{cDown!18}4.67\pct{-9.86\%} & \cellcolor{cDown!18}2.03\pct{-7.58\%} & \cellcolor{cUp!36}0.62\pct{+25.44\%} & \cellcolor{cDown!36}1.22\pct{-26.88\%} & \cellcolor{cUp!10}3.19\pct{+0.84\%} & \cellcolor{cUp!36}0.78\pct{+22.80\%} \\
\midrule
\multirow{6}{*}{DevSkills}
 & Pass@1 (pp) &
 \cellcolor{cUp!10}73.50\pct{-1.50}
 & \cellcolor{cNoise}73.00\pct{+0.50} & \cellcolor{cNoise}73.00\pct{-2.17} & \cellcolor{cNoise}69.83\pct{+2.67} & \cellcolor{cNoise}52.00\pct{-2.33} & \cellcolor{cNoise}52.00\pct{+0.00} & \cellcolor{cNoise}62.00\pct{+2.00} & \cellcolor{cNoise}46.00\pct{+1.00} \\
 & Cost (\$) & \cellcolor{cDown!27}0.473\pct{-13.94\%} & \cellcolor{cDown!46}0.372\pct{-41.73\%} & \cellcolor{cDown!27}0.387\pct{-16.03\%} & \cellcolor{cDown!27}0.103\pct{-11.34\%} & \cellcolor{cNoise}0.894\pct{-1.51\%} & \cellcolor{cDown!46}0.733\pct{-33.00\%} & \cellcolor{cDown!18}0.801\pct{-7.88\%} & \cellcolor{cNoise}0.140\pct{+3.57\%} \\
 & CoP & \cellcolor{cNoise}0.643\pct{-11.83\%} & \cellcolor{cDown!46}0.510\pct{-42.13\%} & \cellcolor{cDown!27}0.530\pct{-13.61\%} & \cellcolor{cDown!27}0.148\pct{-14.83\%} & \cellcolor{cNoise}1.718\pct{+3.58\%} & \cellcolor{cDown!46}1.409\pct{-33.19\%} & \cellcolor{cNoise}1.291\pct{-11.21\%} & \cellcolor{cNoise}0.305\pct{+1.28\%} \\
 \cmidrule(lr){2-10}
 & SubRetrv & \cellcolor{cDown!70}0.35\pct{-64.89\%} & \cellcolor{cDown!57}1.86\pct{-51.52\%} & \cellcolor{cDown!36}4.05\pct{-24.82\%} & \cellcolor{cDown!27}5.08\pct{-17.25\%} & \cellcolor{cDown!57}0.60\pct{-59.78\%} & \cellcolor{cDown!46}1.47\pct{-43.57\%} & \cellcolor{cDown!18}4.02\pct{-8.36\%} & \cellcolor{cUp!27}4.23\pct{+12.92\%} \\
 & SimScrpt & \cellcolor{cDown!57}0.15\pct{-54.93\%} & \cellcolor{cDown!70}1.18\pct{-66.17\%} & \cellcolor{cDown!27}4.02\pct{-12.45\%} & \cellcolor{cDown!10}4.42\pct{-0.49\%} & \cellcolor{cDown!70}0.01\pct{-88.91\%} & \cellcolor{cDown!46}0.55\pct{-38.36\%} & \cellcolor{cUp!10}1.85\pct{+0.36\%} & \cellcolor{cDown!10}1.43\pct{-4.02\%} \\
 & ReTest & \cellcolor{cDown!10}1.47\pct{-4.75\%} & \cellcolor{cDown!46}1.37\pct{-38.57\%} & \cellcolor{cDown!46}3.62\pct{-30.05\%} & \cellcolor{cDown!18}2.04\pct{-7.13\%} & \cellcolor{cUp!57}0.72\pct{+45.68\%} & \cellcolor{cDown!36}1.28\pct{-23.29\%} & \cellcolor{cDown!18}3.00\pct{-5.16\%} & \cellcolor{cUp!10}0.66\pct{+3.71\%} \\
\bottomrule
\end{tabular}
\end{adjustbox}
\end{table*}

\vspace{-5pt}
\section{Mitigating Cost-Inefficient Behaviors in Coding Agents}
\vspace{-5pt}

This section evaluates the effectiveness of the three mitigation approaches in optimizing cost efficiency and the identified cost-inefficient behaviors (Table~\ref{tab:rq234-main-general}). Figure~\ref{fig:waste-vs-cost} further relates changes in behavior-attributed cost to task cost. We first compare the three approaches across all settings, then examine each in depth.

\vspace{-10pt}
\subsection{Cross-Approach Comparison}
\begin{figure}[t]
    \vspace{-6pt}
    \centering
    \includegraphics[width=\textwidth]{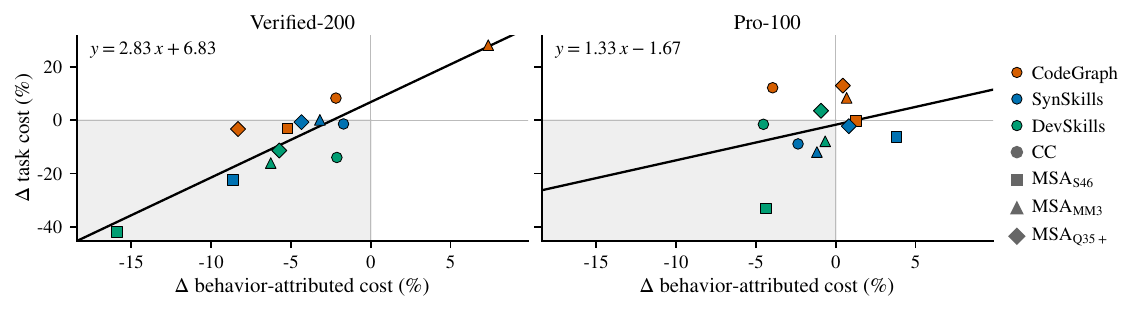}
    \captionsetup{skip=-2pt}
    \caption{Changes in average task cost versus behavior-attributed cost, both relative to baseline task cost. The shaded quadrant indicates joint improvement, and the lines show the fitted linear trends.}
    \label{fig:waste-vs-cost}
    \vspace{-6pt}
\end{figure}

\vspace{-5pt}
\label{sec:cross-approach}
We compare all three approaches across agent configurations and benchmarks. Three patterns emerge.
\textbf{First, mitigation effectiveness generally increases from CodeGraph to agent-synthesized skills to developer-designed skills.} CodeGraph reduces SubRetrv inconsistently and even increases cost robustly in four of eight settings. SynSkills achieve moderate savings, whereas DevSkills produce the broadest reductions in both task cost and diagnosed behaviors.
\textbf{Second, mitigation benefits depend on agent architecture and benchmark.} Improvements are generally larger under \textit{MSA}, while \textit{CC}'s system prompt and tool abstractions already suppress several inefficient behaviors, leaving less room for improvement. All approaches also perform better overall on Verified-200 than Pro-100 in identified behaviors. For skill-based approaches, this partly reflects a generalization challenge, as the skills are derived from Verified trajectories and applied to Pro.
\textbf{Third, task cost changes are amplified relative to the changes in behavior-attributed cost.} In Figure~\ref{fig:waste-vs-cost}, the fitted slopes are 2.83 on Verified-200 and 1.33 on Pro-100. Two effects can amplify the savings. (1) Preventing an inefficient action shortens the trajectory, reducing repeated cache-reading of earlier context in later LLM calls. (2) It can also prevent additional unflagged actions triggered by certain cost-inefficient behaviors. For example, eliminating a ReTest can also avoid subsequent reasoning that interprets its output or diagnoses the same failure.

\subsection{RQ2: Effectiveness of Structure-aware Retrieval}
\vspace{-5pt}
\label{sec:RQ2}

Table~\ref{tab:rq234-main-general} shows that \textbf{CodeGraph neither consistently reduces SubRetrv nor improves task cost efficiency}.\footnote{CodeGraph cannot run on 13 Pro-100 tasks (Appendix~\ref{app:cg-integration}). Its Pro-100 results are therefore measured and compared on the remaining 87 tasks.} Under \textit{CC}, SubRetrv drops by over 75\% on both benchmarks, yet cost rises by 8.30\% on Verified-200 and robustly by 12.19\% on Pro-100. Under \textit{MSA}, SubRetrv increases in four of six settings. Overall, CodeGraph yields no robust cost reduction but four robust increases of 8.39--28.14\%, and Pass@1 remains largely unchanged except for \textit{MSA\textsubscript{Q35+}} on Verified-200 (+5.33 pp). Thus, substantial SubRetrv reduction does not guarantee lower end-to-end cost.

\providecommand{\pct}[1]{\textsubscript{#1}}
\definecolor{cDown}{HTML}{0072B2}
\definecolor{cUp}{HTML}{D2506A}
\begin{table*}[t]
\captionsetup{skip=2pt}
\caption{Effects of CodeGraph on average task token usage, average output tokens per CodeGraph and other retrieval call, and LLM/tool calls. Subscripts show percentage changes from the baselines.}
\label{tab:rq2-mechanism}
\centering
\scriptsize
\setlength{\tabcolsep}{2pt}
\renewcommand{\arraystretch}{0.95}
\begin{adjustbox}{max width=1\textwidth}
\begin{tabular}{l*{8}{c}}
\toprule
& \multicolumn{4}{c}{Verified-200}
& \multicolumn{4}{c}{Pro-100} \\[-2pt]
\cmidrule(lr){2-5}\cmidrule(lr){6-9}
Metric
& \multicolumn{1}{c}{CC}
& \multicolumn{1}{c}{MSA\textsubscript{S46}}
& \multicolumn{1}{c}{MSA\textsubscript{MM3}}
& \multicolumn{1}{c}{MSA\textsubscript{Q35+}}
& \multicolumn{1}{c}{CC}
& \multicolumn{1}{c}{MSA\textsubscript{S46}}
& \multicolumn{1}{c}{MSA\textsubscript{MM3}}
& \multicolumn{1}{c}{MSA\textsubscript{Q35+}} \\[-2pt]
\midrule
Task tokens (M) & \cellcolor{cDown!12}0.91\pct{-1.70\%} & \cellcolor{cDown!12}1.09\pct{-0.37\%} & \cellcolor{cUp!45}4.13\pct{+29.29\%} & \cellcolor{cDown!12}1.63\pct{-1.15\%} & \cellcolor{cDown!12}1.70\pct{-0.36\%} & \cellcolor{cDown!12}2.20\pct{-0.32\%} & \cellcolor{cUp!12}7.28\pct{+9.11\%} & \cellcolor{cUp!28}2.39\pct{+13.39\%} \\
\hline
Output tokens / call & \cellcolor{cUp!62}777\pct{+140.96\%} & \cellcolor{cUp!12}233\pct{+5.98\%} & \cellcolor{cUp!12}251\pct{+5.94\%} & \cellcolor{cDown!12}292\pct{-3.39\%} & \cellcolor{cUp!62}955\pct{+55.74\%} & \cellcolor{cUp!12}354\pct{+9.81\%} & \cellcolor{cDown!12}302\pct{-5.64\%} & \cellcolor{cDown!12}370\pct{-5.56\%} \\
\quad CodeGraph  & 5,199 & 876 & 1,205 & 1,122 & 5,273 & 1,111 & 1,007 & 1,111 \\
\quad Other retrieval & 313 & 258 & 277 & 350 & 643 & 395 & 789 & 541 \\
\hline
LLM calls & \cellcolor{cDown!28}19.91\pct{-19.72\%} & \cellcolor{cDown!12}40.79\pct{-7.22\%} & \cellcolor{cUp!28}95.39\pct{+13.60\%} & \cellcolor{cDown!12}60.15\pct{-2.94\%} & \cellcolor{cDown!28}29.18\pct{-23.21\%} & \cellcolor{cDown!12}56.17\pct{-5.27\%} & \cellcolor{cUp!12}143.91\pct{+8.20\%} & \cellcolor{cUp!12}71.05\pct{+5.68\%} \\
\quad Main agent (S46) & \cellcolor{cDown!12}19.91\pct{-0.42\%} & -- & -- & -- & \cellcolor{cUp!12}29.18\pct{+8.17\%} & -- & -- & -- \\
\quad Subagent (H45) & \cellcolor{cDown!62}0.00\pct{-100.00\%} & -- & -- & -- & \cellcolor{cDown!62}0.00\pct{-100.00\%} & -- & -- & -- \\ \hline
Tool calls & \cellcolor{cDown!45}19.43\pct{-26.61\%} & \cellcolor{cDown!12}42.98\pct{-8.44\%} & \cellcolor{cUp!28}94.66\pct{+12.97\%} & \cellcolor{cDown!12}59.57\pct{-3.87\%} & \cellcolor{cDown!45}32.74\pct{-31.47\%} & \cellcolor{cDown!12}66.94\pct{-5.63\%} & \cellcolor{cUp!12}143.37\pct{+8.07\%} & \cellcolor{cUp!12}71.52\pct{+6.39\%} \\
\quad CodeGraph & 2.23 & 5.07 & 6.75 & 6.31 & 3.60 & 8.13 & 6.33 & 5.65 \\
\quad Other retrieval & \cellcolor{cDown!62}7.16\pct{-53.66\%} & \cellcolor{cDown!45}14.20\pct{-34.13\%} & \cellcolor{cDown!12}35.38\pct{-2.92\%} & \cellcolor{cDown!28}22.30\pct{-23.22\%} & \cellcolor{cDown!62}15.68\pct{-53.68\%} & \cellcolor{cDown!28}35.34\pct{-23.87\%} & \cellcolor{cUp!12}75.01\pct{+3.16\%} & \cellcolor{cDown!12}39.34\pct{-3.44\%} \\
\bottomrule
\end{tabular}
\end{adjustbox}
\end{table*}

Table~\ref{tab:rq2-mechanism} explains why. \textbf{For \textit{CC}, verbose CodeGraph feedback and reduced use of cheaper subagents offset the savings from fewer SubRetrv actions}. CodeGraph cuts LLM and tool calls by 19.72--23.21\% and 26.61--31.47\%, respectively, but each query returns 8.2--16.6$\times$ more tokens than other retrieval actions. Total token usage therefore remains nearly unchanged. Meanwhile, \textit{H45} subagent calls drop from 4.81 and 11.03 per task to zero on both benchmarks, while \textit{S46} main-agent calls change only slightly ($-$0.42\% and +8.17\%). Thus, roughly the same token volume shifts to \textit{S46}, whose token price is 3$\times$ that of \textit{H45}, increasing \textit{CC}'s cost by 8.30\% on Verified-200 and robustly by 12.19\% on Pro-100.

\textbf{For \textit{MSA}, cost rises because agents use CodeGraph additively, stacking CodeGraph queries and their verbose feedback on top of ordinary retrieval rather than replacing it.} For \textit{MSA\textsubscript{MM3}} on both benchmarks and \textit{MSA\textsubscript{Q35+}} on Pro-100, ordinary retrieval remains nearly unchanged while agents add 5.65--6.75 CodeGraph calls per task; these extra calls and their verbose outputs raise total token usage by 9.11--29.29\% and cost robustly by 8.39--28.14\%. In the remaining settings, agents do replace ordinary retrieval with CodeGraph queries: ordinary retrieval drops by over 20\%, but the verbose CodeGraph feedback keeps total token usage nearly unchanged, leaving cost savings indistinguishable from the baseline.

\begin{mdframed}[style=findingbox]
\textbf{Finding 4.} Structure-aware retrieval, as exemplified by CodeGraph, is not inherently cost-efficient. Verbose feedback and tool integration can add retrieval overhead and alter agent delegation, leading to inconsistent SubRetrv reduction and higher cost. It should therefore be evaluated with explicit attention to feedback volume, tool integration, and model orchestration.
\end{mdframed}
\vspace{-10pt}
More details on CodeGraph integration and usage are provided in Appendix~\ref{app:codegraph}.

\vspace{-5pt}
\subsection{RQ3: Agent-Synthesized vs. Developer-Designed Skills}
\vspace{-5pt}
\label{sec:RQ3}
Unlike CodeGraph, which changes how agents access code and targets only retrieval inefficiencies, skills can address broader cost-inefficient behaviors. 
SynSkills are configuration-specific sets of 23--41 operational rules. They target the three cost-inefficient behaviors through concrete, scenario-specific instructions, such as handling failed string substitutions, shell quoting issues, and permission errors.
In contrast, DevSkills are seven general behavioral principles shared by all configurations. They require agents to state a concrete hypothesis before retrieval, reuse available context, and justify necessary re-reads (SubRetrv); to persist and revise scripts instead of creating similar variants (SimScrpt); to understand the test harness, capture test output, and rerun tests only when code changes or new evidence warrant it (ReTest); and to break unproductive loops (general).

\textbf{DevSkills are substantially more effective and consistent than SynSkills}. SynSkills robustly reduce average cost in three of eight settings by 8.86--22.32\%, with no robust cost increase or Pass@1 degradation (Table~\ref{tab:rq234-main-general}). In comparison, DevSkills robustly reduce cost in six settings by 7.88--41.73\%, nearly twice the maximum saving from SynSkills, with only one small Pass@1 drop ($-$1.50\,pp for \textit{CC} on Verified-200). The behavior-level results show the same pattern. For example, DevSkills reduce all three behaviors in \textit{MSA\textsubscript{S46}} on both benchmarks, including SubRetrv by 51.52\% and 43.57\%, compared with 11.38\% and a 2.50\% increase under SynSkills.

\textbf{One possible mechanism is that DevSkills encode higher-level, trace-agnostic guidance, whereas SynSkills retain low-level, trace-specific instructions.} For example, to mitigate SimScrpt, the \textit{MSA\textsubscript{S46}} SynSkills specify when to use ephemeral scripts or overwrite scripts, and how to handle non-ASCII shell content. Such rules depend on specific tasks and repositories, limiting generalization. 
In contrast, DevSkills cover the same cases with one principle: persist and reuse artifacts rather than regenerate similar variants. The principle applies to any task, which helps explain the stronger and more consistent effectiveness of DevSkills and the value of human abstraction in behavioral optimization.

\vspace{-3pt}
\begin{mdframed}[style=findingbox]
\textbf{Finding 5.} Agent-synthesized skills tend to produce low-level, trace-specific guidance, limiting their effectiveness and generality. In contrast, developer-designed skills provide high-level, trace-agnostic guidance and achieve broader cost savings, robustly reducing cost by 7.88--41.73\% in six of eight settings.
\end{mdframed}
\vspace{-10pt}
More analysis and specific skill content for RQ3 are  in Appendix~\ref{app:rq3}.

\vspace{-10pt}
\section{Related Work}
\vspace{-5pt}

\paragraph{Agent Trajectory Analysis.}
Observability platforms support execution tracing and cost monitoring~\citep{langsmith, langfuse, agentops}. Trajectory analyses further examine how agents approach tasks, where execution goes wrong, and how costs accumulate~\citep{lens-traceability, Thought-Action-Result, Understanding-Behaviour, AgentTracer, beyond-final-answer, how-ai-spend-money, Tokenomics}, identifying issues such as repeated actions and loss of coherence~\citep{Graphectory, coherence-collapse}. 
However, how recurring inefficient behaviors contribute to agent costs, why they occur, and whether they can be mitigated remain underexplored. To address this gap, we define and detect three cost-inefficient behaviors and analyze their prevalence, monetary impact, and underlying mechanisms across coding agents.

\vspace{-5pt}
\paragraph{Agent Cost Reduction.}
Existing approaches improve agent cost efficiency through four strategies. 
Context compression and interaction reduction decrease the token usage of each step and the number of interactions~\citep{more-with-less, reduce-code-format, Concise-Thoughts, SWE-Pruner, ContextSniper, llm-step-compression, token-budget-prediction, Compressing-Code-Context}.
Agent orchestration allocates work across suitable models and tools~\citep{understand-sections-with-diff-model, AgentTTS, Toolorchestra, budget-aware, routellm}. 
Runtime supervision intervenes when agents make errors, enter loops, or deviate from expected workflows~\citep{SupervisorAgent, Graphectory}. However, their effects on specific cost-inefficient behaviors remain underexplored. We complement them by using diagnosed behaviors to guide optimizations and evaluating their effects on these behaviors, cost, and task performance.
A fourth strategy improves agent toolsets. Structure-aware retrieval, for example, uses code graphs to guide repository navigation~\citep{RepoGraph, CodexGraph}. Despite reported gains in localization accuracy and efficiency~\citep{LocAgent}, whether these gains consistently translate into end-to-end cost savings remains unclear. We therefore examine its effects on retrieval efficiency and task cost across agent configurations. 

\vspace{-5pt}
\paragraph{Agent Skills.}
Agent skills~\citep{anthropic-skills} provide reusable instructions and optional resources to guide task executions. Recent systems build skills from execution experience~\citep{Trace2Skill, EvoSkill, SkillClaw, AutoSkill, MetaClaw, Skillforge}, incorporating reinforcement learning~\citep{SkillRL, ARISE}, across web, multimodal, and generalist-agent settings~\citep{SkillWeaver, XSkill, Memento-Skills}. Prior work evaluates skill effectiveness~\citep{SkillsBench} and optimizes skills for both task success and inference cost~\citep{Skillmoo}. We derive skills from diagnosed cost-inefficient behaviors and compare our agent-synthesized and developer-designed skills in terms of agent cost efficiency.

\vspace{-10pt}
\section{Threats to Validity}
\vspace{-6pt}
\label{sec:threats}
\textbf{Agent and Task Coverage.}
Budget limits our study to four agent configurations, SWE-bench Verified, and 100 SWE-bench Pro tasks. Behaviors and intervention effects may differ elsewhere.
\textbf{Behavior Identification.}
Parsing, labeling, and detection mechanisms may introduce detection errors despite explicit implementations and manual calibration.
\textbf{Cost and Outcome Measurement.}
We measure LLM inference monetary cost, excluding execution time and hardware costs. Pass@1 may miss regressions beyond test satisfaction.
\textbf{Execution Randomness.}
We triplicate baselines and nine intervention settings; budget limits others to single runs. Robustness estimates rely on baseline-derived variability, which may not fully establish statistical significance.
\textbf{Structure-Aware Retrieval.}
CodeGraph's popularity and robust structural retrieval implementations motivate its selection, but this may limit generalization to other tools.
\textbf{Skill Design and Scope.}
Findings depend on our agent-synthesis pipeline and developer-skill design. Script-based or other skill formats may differ from our prompt-based interventions.

\vspace{-10pt}
\section{Conclusions \& Future Work}
\vspace{-7pt}
\label{sec:conclusion}
We analyze Claude Code and Mini-SWE-Agent trajectories on SWE-bench Verified, identifying three prevalent cost-inefficient behaviors: subsumed retrieval, similar script generation, and test re-execution. Evaluating three mitigation approaches on Verified and Pro shows that structure-aware retrieval can increase cost due to its verbose feedback and altered agent delegation. Agent-synthesized skills offer low-level, trace-specific guidance and moderate savings, while developer-designed skills provide high-level, trace-agnostic guidance and achieve the strongest and most consistent savings. Future work will extend this analysis to longer-horizon, interactive tasks and more diverse agents, examining inefficiencies beyond monetary cost. One could also use behavior diagnoses to guide optimizations and generate more efficient trajectories for agent training.

\section{AI Use Statement}
\label{sec:ai-usage}
\vspace{-5pt}
\paragraph{AI within our experimental methods.}
Analyzing and mitigating the cost-inefficient behaviors of the coding agents is the object of our study, and their benchmark execution trajectories are used as our experimental data. We combine rule-based parsing with LLM-based action labeling under an author-defined taxonomy (Section~\ref{sec:study-setup}). The authors design the labeling rules and thresholds, manually calibrate them, and examine their application to agent trajectories. In RQ3 (Section~\ref{sec:RQ3}), an LLM-based pipeline generates agent-synthesized skills as a mitigating approach for evaluation.

\vspace{-0.15in}
\paragraph{AI assistance in research and writing.}
We use generative AI assistants to help debug and refactor author-written code, refine manuscript structure and wording, and improve table and figure layouts. The authors independently design the research questions, metrics, detection rules, analysis, findings, and conclusions. We review the AI-assisted work and take responsibility for the final manuscript, code, analysis, and results.

\ificlrfinal
\section*{Acknowledgments}
This research was supported in part by NSF 1901242 and 2006688. Any opinions, findings, and conclusions in this paper are those of the authors only and do not necessarily reflect the views of our sponsors.
\fi

\bibliographystyle{iclr2027_conference}
\bibliography{references}

\appendix
\newpage
\section{Appendix}
\label{sec:appendix}
This appendix provides supplementary details for our study.
First, we describe additional study setup details, including the versions and configurations of Claude Code and Mini-SWE-Agent, action-labeling procedures, benchmark task split, and how we handle the agent execution randomness.
Second, we provide additional results and analysis for RQ1, including the performance of the four agent configurations on the 300-task RQ1 analysis set, extra detection details, analysis, and examples of the three cost-inefficient behaviors.
Third, we provide CodeGraph integration details and further results analysis for RQ2.
Finally, we detail the implementation of agent-synthesized skills and provide the full content of both agent-synthesized and developer-designed skills in RQ3.

\vspace{-6pt}
\subsection{Additional Details for Study Setup}
\vspace{-6pt}
\label{app:implementation-detail}

In this section, we will provide the specific version and configuration details of the agents included in our study, and then we will introduce the agent action labeling process, benchmark task split, and the details of handling agent execution randomness.

We use the same agent versions and configurations across all RQs. \textit{CC} uses CLI v2.1.169~\citep{claude-code-npm} with a \$3.00 per-task cost limit for SWE-bench Verified tasks and a \$5.00 per-task cost limit for SWE-bench Pro tasks; the backbone model temperature is not exposed through the CLI. \textit{MSA} uses v2.3.0 with temperature 0.0, a 250-step limit, and the same per-task cost limit as CC. We next elaborate on the action-labeling procedure, benchmark split details, and our treatment of execution randomness described in Section~\ref{sec:study-setup}.

\newcommand{\codenew}[1]{\lstinline[basicstyle=\ttfamily\scriptsize,breaklines=true]@#1@}

\begin{table}[t]
\centering
\scriptsize
\captionsetup{skip=3pt}
\caption{Action-label taxonomy.}
\label{tab:action-labels}
\setlength{\tabcolsep}{4pt}
\renewcommand{\arraystretch}{0.9}
\begin{tabularx}{0.9\textwidth}{@{}l l X@{}}
\toprule
\textbf{Category} & \textbf{Label} & \textbf{Description} \\
\midrule
\multirow{4}{*}{Retrieval}
 & \codenew{Retrieve\_Structure} & Searches or explores the repository structure or current location. \\
 & \codenew{Retrieve\_Code} & Reads or searches source code files, except test files or documents. \\
 & \codenew{Retrieve\_Tests} & Reads or searches test files specifically. \\
 & \codenew{Retrieve\_Docs} & Reads documentation (non-code references such as README). \\
\midrule
\multirow{5}{*}{Modification}
 & \codenew{Generate\_Patch} & Modifies a file inside the working tree (e.g., \codenew{/testbed}) to fix the issue. \\
 & \codenew{Generate\_Tests} & Writes tests or reproduction scripts (e.g., \codenew{python -c}). \\
 & \codenew{Generate\_Tools} & Creates custom tools for the agent's own use. \\
 & \codenew{Create\_Files\_Directories} & Creates new files or directories (e.g., \codenew{touch}, \codenew{mkdir}). \\
 & \codenew{Remove\_Content} & Removes code or files the agent previously created. \\
\midrule
\multirow{3}{*}{Verification}
 & \codenew{Execute\_Existing\_Tests} & Runs existing repository tests. \\
 & \codenew{Execute\_Generated\_Tests} & Executes agent-generated code. \\
 & \codenew{Review\_Patch} & Reviews the applied patch (e.g., \codenew{git diff}). \\
\midrule
Git & \codenew{Execute\_Git} & Runs git commands other than \codenew{git diff}. \\
\midrule
\multirow{6}{*}{Other}
 & \codenew{Finish\_Task} & Submits the final solution. \\
 & \codenew{Env\_Inspection} & Checks or installs packages; inspects environment variables or resources. \\
 & \codenew{Pure\_Reasoning} & The step contains only reasoning, without calling any tool. \\
 & \codenew{Delegate\_Subagent} & Dispatches a subagent via the \codenew{Agent} tool (\textit{CC} only). \\
 & \codenew{Other} & Does not fit any label above. \\
\bottomrule
\end{tabularx}
\end{table}

\vspace{-6pt}
\subsubsection{Action Labeling}
\vspace{-6pt}
\label{app:labeling}
\textbf{Why label actions.}
The action labels serve four purposes. 
First, the two agent frameworks expose actions in different surface forms: \textit{CC} issues named tool calls while \textit{MSA} issues raw shell commands. We therefore assign every step one or two semantic action labels before any detection, which projects both frameworks onto a shared vocabulary and makes their trajectories comparable. 
Second, they restrict each detector to its relevant step population: SubRetrv compares returned context only across retrieval steps, SimScrpt mines scripts only from script-generation or script-execution steps, and ReTest groups only test-execution steps. 
Third, they provide the structural boundaries the detections rely on. For example, a step labeled \code{Generate\_Patch} closes the current ReTest group, so re-running a test after an updated patch is never flagged. 
Finally, they support the aggregate analysis, such as the retrieval-action accounting in Table~\ref{tab:redundant-retrieval-reasons} (Patch-Adjacent SubRetrv and Within-Sequence SubRetrv). 
The same labeling pipeline is applied uniformly to every run in RQ1--RQ3.

\textbf{Label taxonomy.}
We use 18 labels in five categories, summarized in Table~\ref{tab:action-labels}. Action \code{Delegate\_Subagent} arises only in \textit{CC}'s tool set. A tool call may carry up to two labels. For example, an inline \code{python -c} reproduction script both authors and executes agent-generated code (\code{Generate\_Tests} \& \code{Execute\_Generated\_Tests}).

\textbf{Rule-based labeling.}
Steps are first labeled by deterministic rules, with one rule family per harness. For \textit{MSA}, whose actions are raw shell commands, each command is split into atomic sub-commands (\code{\&\&}, \code{;}, and pipes) after unwrapping prefixes such as \code{timeout}; navigation-only fragments (\code{cd}, \code{export}) are dropped. Read-only verbs (\code{cat}, \code{head}, \code{sed -n}, \code{grep}, \code{ls}, \code{find}, \ldots) map to \emph{Retrieval}, with test-path, documentation-path, and source-extension patterns selecting the sub-label. Write channels (heredocs, \code{sed -i}, \code{tee}, output redirection) map to \emph{Modification}. Modifying an existing file inside the working directory yields \code{Generate\_Patch}; a test-looking file yields \code{Generate\_Tests}; ephemeral script generations (\code{python -c}, \code{python <<EOF}) are labeled as \code{Generate\_Tests} $+$ \code{Execute\_Generated\_Tests}, unless they call an existing test API or open a working-tree file for writing. Test runners map to \code{Execute\_Existing\_Tests} or \code{Execute\_Generated\_Tests} depending on their target; \code{git diff} maps to \code{Review\_Patch}; other \code{git} commands map to \code{Execute\_Git}; the submission marker (MSA only) maps to \code{Finish\_Task}; and steps without any command map to \code{Pure\_Reasoning}. For \textit{CC}, whose actions are named tool calls, tools map directly (\code{Read}/\code{Grep} to \emph{Retrieval} by path, \code{Glob} to \code{Retrieve\_Structure}, \code{Edit}/\code{Write} to \emph{Modification} by path, \code{Agent} to \code{Delegate\_Subagent}), and \code{Bash} calls are routed through the same command classifier as \textit{MSA}.

\textbf{LLM fallback.}
A step is forwarded to an LLM judge (GPT-5.1) whenever any of its atomic commands matches no rule. The judge receives the step's thought and commands together with the taxonomy definitions and returns at most two labels with a brief justification. Among the RQ1 trajectories, the fallback handles 0.94\% of \textit{CC} steps, 31.17\% of \textit{MSA\textsubscript{S46}} steps, 1.62\% of \textit{MSA\textsubscript{MM3}} steps, and 1.18\% of \textit{MSA\textsubscript{Q35+}} steps, accounting for 7.26\% of the total steps of four agent configurations on 300 Verified tasks. By submitting all LLM-labeling requests through the batch mode, the labeling cost is less than 0.2 cents per trajectory.

\begin{figure*}[t]
    \centering
    \includegraphics[width=0.9\textwidth]{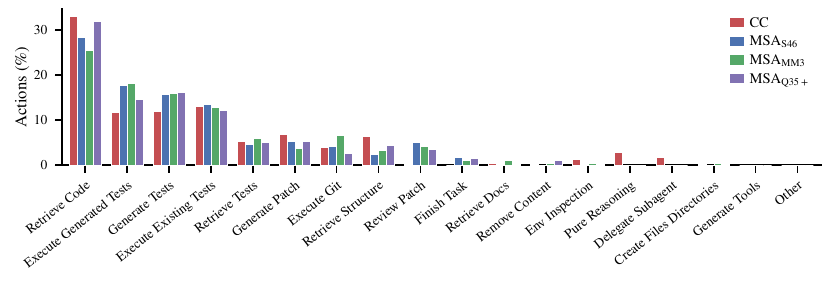}
    \captionsetup{skip=-6pt}
    \caption{Distribution of action labels over the RQ1 trajectories.}
    \label{fig:label-distribution}
\end{figure*}

\textbf{Resulting distribution.}
Figure~\ref{fig:label-distribution} shows the label distribution over the RQ1 trajectories among four agent configurations. The pattern is consistent across configurations: reading source code (\code{Retrieve\_Code}, 25.65--33.12\%) together with generating and executing scripts (\code{Generate\_Tests} $+$ \code{Execute\_Generated\_Tests}, 23.66--34.07\%) dominate every configuration, followed by existing-test execution (12.15--13.57\%). Architecture differences appear in the tail: only \textit{CC} delegates work to subagents (\code{Delegate\_Subagent}, 1.83\% of its actions), while \textit{MSA} configurations review patches through \code{git diff} more often (3.57--5.04\% vs.\ 0.32\%).

\begin{table*}[t]
\centering
\caption{Task selection by repository on SWE-bench Verified and SWE-bench Pro.}
\label{tab:task-selection}
\scriptsize
\renewcommand{\arraystretch}{1}
\setlength{\tabcolsep}{6pt}
\begin{tabular}{|l|r|r||l|r|r|}
\hline
\multicolumn{3}{|c||}{\textbf{SWE-bench Verified} (500 tasks)} &
\multicolumn{3}{c|}{\textbf{SWE-bench Pro} (731 public tasks)} \\
\hline
\textbf{Repository} & \textbf{Analysis} & \textbf{Eval.} &
\textbf{Repository} & \textbf{Pool} & \textbf{Eval.} \\
\hline
astropy/astropy           & 13  & 9  & ansible/ansible             & 96 & 12 \\
django/django             & 139 & 92 & element-hq/element-web      & 56 & 8  \\
matplotlib/matplotlib     & 20  & 14 & flipt-io/flipt              & 85 & 10 \\
mwaskom/seaborn           & 1   & 1  & future-architect/vuls       & 62 & 9  \\
pallets/flask             & 1   & 0  & gravitational/teleport      & 76 & 10 \\
psf/requests              & 5   & 3  & internetarchive/openlibrary & 91 & 11 \\
pydata/xarray             & 13  & 9  & navidrome/navidrome         & 57 & 8  \\
pylint-dev/pylint         & 6   & 4  & NodeBB/NodeBB               & 44 & 8  \\
pytest-dev/pytest         & 12  & 7  & protonmail/webclients       & 65 & 9  \\
scikit-learn/scikit-learn & 19  & 13 & qutebrowser/qutebrowser     & 79 & 10 \\
sphinx-doc/sphinx         & 26  & 18 & tutao/tutanota              & 20 & 5  \\
sympy/sympy               & 45  & 30 &                             &    &    \\
\hline
\textbf{Total} & \textbf{300} & \textbf{200} &
\textbf{Total} & \textbf{731} & \textbf{100} \\
\hline
\end{tabular}
\end{table*}

\subsubsection{Behavior Cost Attribution}
\label{app:behavior-cost}

For each action flagged as cost-inefficient, we use the \texttt{o200k\_base} tokenizer from \texttt{tiktoken} to count the tokens in its reasoning, commands, and tool output. We then attribute costs using the corresponding model's token prices. Generated content is charged at the output-token price, while both generated content and tool output incur input-token or cache-token charges when carried into subsequent LLM requests. We sum these charges to obtain the action's attributed cost.

\vspace{-6pt}
\subsubsection{Benchmarks \& Task Split}
\label{app:data-select}
Table~\ref{tab:task-selection} lists the tasks per repository used for analysis in \textbf{RQ1} and for evaluation in \textbf{RQ2--RQ3}. For \emph{SWE-bench Verified}, we order the tasks of each repository by creation date and assign the earliest tasks, about 60\% per repository, to the analysis set and the remaining ones to \emph{Verified-200}. \code{pallets/flask} has a single task, which falls into the analysis set. 

For \emph{SWE-bench Pro}, we draw a stratified random sample of 100 tasks from the 731 public tasks with a fixed seed. The \emph{Pool} column gives the public tasks available in each repository. Each repository receives a quota that grows with the square root of its pool size, with a minimum of four tasks, so that all 11 repositories are represented, and repositories with more tasks do not crowd out smaller ones. Within each repository, tasks are ranked by their cost in the public SWE-bench Pro leaderboard run of SWE-agent~\citep{swe-agent} with Claude Sonnet 4.5 (250-turn limit, no cost cap)\footnote{When we selected the tasks on June 26, 2026, this was the best-performing leaderboard run whose released trajectories include cost information (43.70\% resolved). The share of resolved tasks in \emph{Pro-100} matches that of this run, with 44 passed tasks.} and split into three equal-sized terciles. The low, mid, and high terciles are sampled at 20\%, 30\%, and 50\%, since costlier tasks are expected to expose more cost-inefficient behaviors.

\begin{table*}[t]
\centering
\caption{Noise floor of the eight baselines. The floor $s$ is the standard deviation of Pass@1 (pp) and the coefficient of variation of cost and CoP (\%). The last two columns give the smallest single-run $|\Delta|$ classified as robust ($1.90\,s$). Per-run cost is averaged over the tasks completed in all three runs.}
\label{tab:app-noise-floors}
\scriptsize
\setlength{\tabcolsep}{3pt}
\renewcommand{\arraystretch}{0.9}
\begin{tabular}{@{}ll c r c rr rr@{}}
\toprule
& & \multicolumn{2}{c}{\textbf{Pass@1}} & \multicolumn{2}{c}{\textbf{Cost}} & \textbf{CoP} & \multicolumn{2}{c}{\textbf{Min.\ robust $|\Delta|$, $R{=}1$}} \\
\cmidrule(lr){3-4}\cmidrule(lr){5-6}\cmidrule(lr){7-7}\cmidrule(l){8-9}
\textbf{Benchmark} & \textbf{Config} & Run 1 / 2 / 3 (\%) & $s_{\mathrm{pass}}$ (pp) & Run 1 / 2 / 3 (\$) & $s_{\mathrm{cost}}$ (\%) & $s_{\mathrm{CoP}}$ (\%) & Cost & Pass@1 (pp) \\
\midrule
Verified-200 & CC & 74.00 / 75.50 / 75.50 & 0.87 & 0.598 / 0.525 / 0.515 & 8.35 & 9.56 & 15.87\% & 1.65 \\
Verified-200 & MSA\textsubscript{S46} & 72.00 / 71.50 / 74.00 & 1.32 & 0.674 / 0.605 / 0.636 & 5.40 & 5.49 & 10.25\% & 2.51 \\
Verified-200 & MSA\textsubscript{MM3} & 78.00 / 71.50 / 76.00 & 3.33 & 0.463 / 0.442 / 0.477 & 3.80 & 2.85 & 7.21\% & 6.33 \\
Verified-200 & MSA\textsubscript{Q35+} & 65.00 / 70.50 / 66.00 & 2.93 & 0.123 / 0.110 / 0.115 & 5.70 & 9.68 & 10.82\% & 5.57 \\
\midrule
Pro-100 & CC & 55.00 / 56.00 / 52.00 & 2.08 & 0.917 / 0.922 / 0.881 & 2.44 & 1.94 & 4.64\% & 3.95 \\
Pro-100 & MSA\textsubscript{S46} & 55.00 / 49.00 / 52.00 & 3.00 & 1.037 / 1.067 / 0.983 & 4.14 & 8.43 & 7.87\% & 5.70 \\
Pro-100 & MSA\textsubscript{MM3} & 60.00 / 64.00 / 56.00 & 4.00 & 0.859 / 0.852 / 0.896 & 2.70 & 9.32 & 5.13\% & 7.60 \\
Pro-100 & MSA\textsubscript{Q35+} & 47.00 / 41.00 / 47.00 & 3.46 & 0.126 / 0.126 / 0.132 & 2.83 & 7.22 & 5.38\% & 6.58 \\
\bottomrule
\end{tabular}
\end{table*}

\vspace{-6pt}
\subsubsection{Controlling Agent Execution Randomness}
\label{app:randomness}

This section describes how the run-to-run noise floors in \textbf{RQ2--RQ3} are measured, how the effect of single-run approach cells is judged, and how the underlying equal-variance assumption was validated. All numbers are produced by the same pipeline that generates Table~\ref{tab:rq234-main-general}.

\textbf{Design Overview.}
Re-running every (benchmark, configuration, approach, and baseline) cell three times (2 $\times$ 4 $\times$ $(3+1)$ $\times$ 3 $\ =\ 96$ runs in total) would cost about \$7,500 at the per-run costs we observed, so we proceed in three steps with selective re-running 8 baseline settings and 9 approach cells. First, every \emph{baseline} setting (two benchmarks $\times$ four agent configurations) is executed three times to estimate the run-to-run noise floors in Table~\ref{tab:app-noise-floors}, following standard practice in performance benchmarking~\citep{java-performance-eval, rigorous-benchmarking} and LLM evaluations~\citep{eval-error-bars}. 
Second, 15 approach cells are executed once, and their effects are judged against the noise floors transferred from their corresponding baseline, \emph{under the equal-variance assumption that our approaches do not materially alter the sampling and infrastructure nondeterminism driving run-to-run noise}. 
We further verify this assumption directly on the nine replicated cells in Table~\ref{tab:app-repeats}, selected to cover both benchmarks, all three approaches, and both agent frameworks at low replication cost. Here are the nine approach cells we select to replicate: six cells run all three approaches on both benchmarks with \textit{MSA\textsubscript{Q35+}}, the cheapest configuration to replicate; one cell runs developer-designed skills on Verified-200 with \textit{MSA\textsubscript{MM3}}, the noisiest configuration based on its baseline noise floor; the remaining two cells, whose single-run effects are borderline, run developer-designed skills and CodeGraph on Verified-200 with \textit{CC}. 
In total, the \textbf{RQ2--RQ3} evaluation comprises over 10k trajectories in SWE-bench Verified and Pro. For RQ1, each agent configuration is only executed once on 300 Verified tasks.
The remainder of this section details these steps in turn: the measured noise floors, the decision rule that classifies an observed effect as robust, the validation of the equal-variance assumption together with a stress test showing that the verdicts hold even if it were violated, and the reason behavior metrics carry no robustness verdicts.

\begin{table*}[t]
\centering
\caption{Approach cells executed three times to verify the equal-variance assumption. Each metric lists the approach and baseline noise ($s_{\mathrm{appr}}$ / $s_{\mathrm{base}}$) and their ratio. The remaining 15 approach cells are single runs.}
\vspace{-10pt}
\label{tab:app-repeats}
\scriptsize
\setlength{\tabcolsep}{2.5pt}
\renewcommand{\arraystretch}{0.9}
\resizebox{\textwidth}{!}{%
\begin{tabular}{@{}lll c cr c cr r@{}}
\toprule
& & & \multicolumn{3}{c}{\textbf{Pass@1 (pp)}} & \multicolumn{3}{c}{\textbf{Cost (\$)}} & \textbf{CoP} \\
\cmidrule(lr){4-6}\cmidrule(lr){7-9}\cmidrule(l){10-10}
\textbf{Benchmark} & \textbf{Config} & \textbf{Approach} & Per run & $s_{\mathrm{appr}}$ / $s_{\mathrm{base}}$ & Ratio & Per run & $s_{\mathrm{appr}}$ / $s_{\mathrm{base}}$ & Ratio & Ratio \\
\midrule
Verified-200 & CC & DevSkills & 73.00 / 73.50 / 74.00 & 0.50 / 0.87 & 0.58 & 0.497 / 0.487 / 0.435 & 7.06 / 8.35 & 0.84 & 0.80 \\
Verified-200 & CC & CodeGraph & 73.50 / 73.50 / 75.50 & 1.15 / 0.87 & 1.33 & 0.590 / 0.613 / 0.587 & 2.36 / 8.35 & 0.28 & 0.36 \\
Verified-200 & MSA\textsubscript{MM3} & DevSkills & 72.00 / 72.50 / 74.50 & 1.32 / 3.33 & 0.40 & 0.371 / 0.381 / 0.410 & 5.23 / 3.80 & 1.38 & 1.19 \\
Verified-200 & MSA\textsubscript{Q35+} & DevSkills & 67.50 / 72.00 / 70.00 & 2.25 / 2.93 & 0.77 & 0.107 / 0.099 / 0.103 & 3.83 / 5.70 & 0.67 & 0.73 \\
Verified-200 & MSA\textsubscript{Q35+} & SynSkills & 70.00 / 66.00 / 73.00 & 3.51 / 2.93 & 1.20 & 0.119 / 0.112 / 0.115 & 3.01 / 5.70 & 0.53 & 0.48 \\
Verified-200 & MSA\textsubscript{Q35+} & CodeGraph & 73.00 / 72.00 / 72.50 & 0.50 / 2.93 & 0.17 & 0.109 / 0.115 / 0.113 & 3.01 / 5.70 & 0.53 & 0.38 \\
Pro-100 & MSA\textsubscript{Q35+} & DevSkills & 48.00 / 42.00 / 48.00 & 3.46 / 3.46 & 1.00 & 0.140 / 0.139 / 0.142 & 1.38 / 2.83 & 0.49 & 0.97 \\
Pro-100 & MSA\textsubscript{Q35+} & SynSkills & 49.00 / 45.00 / 44.00 & 2.65 / 3.46 & 0.76 & 0.130 / 0.132 / 0.133 & 1.33 / 2.83 & 0.47 & 1.03 \\
Pro-100 & MSA\textsubscript{Q35+} & CodeGraph & 41.38 / 35.63 / 40.23 & 3.04 / 4.14 & 0.73 & 0.146 / 0.164 / 0.141 & 8.19 / 3.43 & 2.39 & 1.51 \\
\bottomrule
\end{tabular}}
\end{table*}

\textbf{Noise floors.}
Table~\ref{tab:app-noise-floors} lists the three baseline runs of eight settings. The noise floor $s$ is the standard deviation (SD) of Pass@1 (in pp), and the coefficients of variation (CV) of cost (in \%) and CoP (in \%). Pass@1 floors range from 0.87\,pp (\textit{CC} on Verified-200) to 4.00\,pp (\textit{MSA\textsubscript{MM3}} on Pro-100). The stability order within each benchmark is \textit{CC} $>$ \textit{MSA\textsubscript{S46}} $>$ \textit{MSA\textsubscript{Q35+}} $>$ \textit{MSA\textsubscript{MM3}}. Cost floors range from 2.44\% to 8.35\%. CoP floors are generally wider than cost floors (up to 3.45$\times$, \textit{MSA\textsubscript{MM3}} on Pro-100) because CoP inherits the variance of the Pass@1 in its denominator. For CodeGraph on Pro-100, which runs on 87 tasks (Appendix~\ref{app:cg-integration}), all noise floors are re-estimated on the same 87 tasks: 3.51, 3.51, 3.45, and 4.14\,pp for Pass@1, 2.59\%, 3.99\%, 2.67\%, and 3.43\% for cost, and 5.91\%, 9.20\%, 8.28\%, and 10.81\% for CoP, for \textit{CC}, \textit{MSA\textsubscript{S46}}, \textit{MSA\textsubscript{MM3}}, and \textit{MSA\textsubscript{Q35+}}, respectively.

\textbf{Robust decision rule.}
Cost and CoP changes ($\Delta$) are computed on the tasks completed in every run of both the approach and its baseline. For an approach cell with $R$ runs compared to its three-run baseline, the standard error of $\Delta$ is $\sigma_\Delta = s\sqrt{1/R + 1/3}$. We classify $\Delta$ as robust when $|\Delta|/\sigma_\Delta \ge 1.645$, i.e., when a replication would reproduce its sign with probability at least 95\% under a normal model~\citep{replication, reproducibility}. \textbf{For a single-run cell this reduces to $|\Delta| \ge 1.90\,s$, which is a stricter bar than a replicated cell's}. The resulting thresholds for each setting are given in the last two columns of Table~\ref{tab:app-noise-floors}.

\textbf{Validating the equal-variance assumption.}
Table~\ref{tab:app-repeats} reports nine approach cells executed three times. All other 15 approach cells are single runs. Across the nine replicated cells, the approach-side Pass@1 SD is 0.17--1.33$\times$ the baseline's, and the cost SD is 0.28--1.38$\times$ the baseline's in eight of the nine cells. However, these ratios are themselves imprecise: for three normal runs, the sample variance satisfies $2s^2/\sigma^2 \sim \chi^2_2$, so an estimated $s$ carries a relative standard error of $\sqrt{1-\pi/4}\,/\,(\sqrt{\pi}/2) \approx 52\%$, and ratios moderately above one are expected even under equal noise. Approach runs are thus not systematically noisier than baseline runs, which is what the transfer of baseline noise floors requires. The single exception is CodeGraph with \textit{MSA\textsubscript{Q35+}} on Pro-100, whose cost noise is 2.39$\times$ its baseline's (8.19\% vs.\ 3.43\%). For this cell we use its own observed noise floor (8.19\% for cost and 16.34\% for CoP). Its cost and CoP increases remain robust. The same amplification is not observed for CodeGraph with \textit{CC} on Verified-200 (cost ratio 0.28), so we do not widen CodeGraph floors elsewhere. 

Moreover, the verdicts in Table~\ref{tab:rq234-main-general} carry a safety margin beyond this validation: even if the equal-variance assumption were violated at the largest magnitude we observed outside the exception, none of these verdicts would change for two reasons. First, every single-run effect classified as robust remains robust even after its transferred noise floor is inflated by $1.5\times$, whereas the largest amplification observed among the replicated cells outside the exception is only $1.38\times$ ( the cost of DevSkills with \textit{MSA\textsubscript{MM3}} on Verified-200, in Table~\ref{tab:app-repeats}). Second, though each robust effect retains a residual probability of reversing its sign in a replication, summing these probabilities over the 13 robust cost effects yields only 0.06 expected sign reversals if the entire study were rerun.

\textbf{Behavior metrics.}
The per-task behavior frequencies reported in Table~\ref{tab:rq234-main-general} are considerably noisier than cost. Across the eight baselines, the CV of a behavior's per-task frequency over the three runs averages 15.68\% and reaches 44.47\% (SimScrpt of \textit{MSA\textsubscript{Q35+}} on Pro-100), whereas cost floors range from 2.44\% to 8.35\%; it exceeds the cost floor of the same setting in 20 of the 24 baseline-behavior combinations. We therefore report behavior changes in Table~\ref{tab:rq234-main-general} as mechanistic evidence and do not attach robustness verdicts to them.

\vspace{-10pt}
\subsection{Additional Results for RQ1}
\vspace{-10pt}
\label{app:rq1}

In this section, we provide additional results and analysis for RQ1. We first report Pass@1, task cost, and CoP for the four agent configurations on the 300 RQ1 tasks. We further provide additional statistics, analysis, and illustrative examples for the three identified cost-inefficient behaviors.

\begin{wraptable}{r}{0.5\textwidth}
\vspace{-12pt}
\centering
\scriptsize
\captionsetup{
    width=0.95\linewidth,
    justification=raggedright,
    singlelinecheck=false
}
\captionsetup{skip=3pt}
\caption{Pass@1, cost, and CoP of the four agent configurations in \textbf{RQ1}.}
\label{tab:app-baseline300}
\setlength{\tabcolsep}{4pt}
\renewcommand{\arraystretch}{0.9}
\begin{tabular}{@{}l rrr@{}}
\toprule
\textbf{Config} & \textbf{Pass@1} & \textbf{Cost (\$)} & \textbf{CoP} \\
\midrule
CC & 76.67\% & 0.520 & 0.679 \\
MSA\textsubscript{S46} & 74.33\% & 0.554 & 0.745 \\
MSA\textsubscript{MM3} & 72.67\% & 0.365 & 0.502 \\
MSA\textsubscript{Q35+} & 65.67\% & 0.102 & 0.156 \\
\bottomrule
\end{tabular}
\vspace{-8pt}
\end{wraptable}

\textbf{Task performance and cost.} Table~\ref{tab:app-baseline300} reports Pass@1, average task cost, and CoP of the four agent configurations on the RQ1 analysis set. \textit{CC} achieves the highest Pass@1 (76.67\%) at a per-task cost slightly below \textit{MSA\textsubscript{S46}} (\$0.52 vs.\ \$0.55). \textit{MSA\textsubscript{MM3}} trades 1.67\,pp of Pass@1 against \textit{MSA\textsubscript{S46}} for a 34.04\% lower task cost, and \textit{MSA\textsubscript{Q35+}} is the cheapest configuration (\$0.10 per task) but also the least accurate (65.67\%).

\subsubsection{Why CC Exhibits Fewer Cost-Inefficient Behaviors than MSA}
\label{app:rq1-cc-better}

A large part of \textit{CC}'s advantage lies in its built-in system prompt and tool descriptions~\citep{claude-code-npm, cc-system-prompts}. They already contain instructions that mitigate inefficient behaviors before any external optimization approach is applied, as listed below.

\textbf{SubRetrv:}
\begin{itemize}[left=0pt, itemsep=2pt, topsep=2pt, parsep=0pt, partopsep=0pt]
\item \code{Bash} tool description: ``IMPORTANT: Avoid using this tool to run commands, unless \ldots a dedicated tool cannot accomplish your task. Instead, use the appropriate dedicated tool\ldots'', with an explicit mapping: file search via \code{Glob} instead of \code{find}, content search via \code{Grep} instead of \code{grep}, file reading via \code{Read} instead of \code{cat}.
\item \code{Agent} tool (used for delegating subagents) description: ``Launch up to N agents IN PARALLEL (single message, multiple tool calls) to efficiently explore the codebase. Use 1 agent when the task is isolated to known files\ldots'' However, this is double-edged: it keeps the main context compact, but the summaries omit the exact code, so the main agent sometimes re-reads the same region later. This causes the Cross-Agent SubRetrv that accounts for 50.15\% of \textit{CC}'s SubRetrv (Table~\ref{tab:redundant-retrieval-reasons}).
\end{itemize}
However, no built-in instructions contain context like ``do not re-read what you already retrieved'', which is one of the reasons why \textit{CC} still has SubRetrv, and other mitigation approaches can still optimize such behavior.

\textbf{SimScrpt:}
\begin{itemize}[left=0pt, itemsep=2pt, topsep=2pt, parsep=0pt, partopsep=0pt]
\item System prompt: ``NEVER create files unless they're absolutely necessary for achieving your goal. ALWAYS prefer editing an existing file to creating a new one.'' It suppresses scratch file creation at the source, making file-based SimScrpt rare on \textit{CC}. \textit{CC} creates 0.21 draft files per task versus 2.72 to 7.07 for \textit{MSA}, and file-based SimScrpt appears in only two \textit{CC} tasks.
\item \code{Edit} tool description: ``ALWAYS prefer editing existing files. NEVER write new files unless explicitly required.'' The rich feedback of the \code{Edit} tool and such descriptions make CC barely generate scripts for code editing, which is a major category in \textit{MSA}'s file-based SimScrpt.
\end{itemize}
Since creating an issue reproduction script can still help solve the task, \textit{CC} achieves this with inline \code{python -c} one-shots. This explains why its SimScrpt is almost entirely ephemeral rather than file-based.

\textbf{ReTest:} we find no built-in instruction that discourages re-running tests without updated patches or asks the agent to reuse previous test output. Consistently, ReTest is the one behavior where \textit{CC} shows no clear advantage in frequency: 2.09 executions per task, the same magnitude as \textit{MSA\textsubscript{S46}} (2.51) and \textit{MSA\textsubscript{Q35+}} (2.10).

\textbf{Other instructions improve cost efficiency but are not tied to the specific cost-inefficient behaviors we identified.} ``You have the capability to call multiple tools in a single response. When multiple independent pieces of information are requested and all commands are likely to succeed, run multiple tool calls in parallel for optimal performance.'' helps reduce round trips; ``Your responses should be short and concise.'' reduces output tokens.

\vspace{-5pt}
\subsubsection{Subsumed Retrieval}
\vspace{-5pt}

Our detection mechanisms for SubRetrv do not consider code clones beyond whitespace differences, which may underestimate SubRetrv; we leave clone-aware detection to future work.

\begin{wraptable}{r}{0.52\textwidth}
\vspace{-12pt}
\centering
\scriptsize
\setlength{\tabcolsep}{2pt}
\renewcommand{\arraystretch}{0.9}
\captionsetup{
    width=0.95\linewidth,
    justification=raggedright,
    singlelinecheck=false
}
\captionsetup{skip=3pt}
\caption{Total retrieved context and the lines re-read by SubRetrv actions.}
\label{tab:redundant-retrieval-lines}
\begin{tabular}{@{}l rrrr@{}}
\toprule
\textbf{Metric} & \textbf{CC} & \textbf{MSA\textsubscript{S46}} & \textbf{MSA\textsubscript{MM3}} & \textbf{MSA\textsubscript{Q35+}} \\
\midrule
Avg.\ total lines read        & 1{,}032 & 573 & 912 & 1{,}051 \\
Avg.\ lines re-read by SubRetrv     & 114 & 96 & 137 & 190 \\
Overlapping rate               & 11.05\% & 16.69\% & 15.00\% & 18.10\% \\
\bottomrule
\end{tabular}
\end{wraptable}

Table~\ref{tab:redundant-retrieval-lines} quantifies the retrieved context behind SubRetrv: across configurations, agents re-read 11.05--18.10\% of all retrieved code lines through SubRetrv actions. Notably, retrieval volume alone does not determine SubRetrv severity. \textit{MSA\textsubscript{S46}} reads the fewest lines per task (573) yet re-reads a larger share (16.69\%) than \textit{CC}, which reads almost twice as many (1,032) with the lowest share (11.05\%). How the architecture manages the retrieved context therefore matters more than how much it retrieves. Below, we illustrate SubRetrv occurrence scenarios with concrete examples.

\Needspace*{10\baselineskip}
\begin{wrapfigure}{r}{0.55\textwidth}
\vspace{-12pt}
\centering
\includegraphics[width=\linewidth]{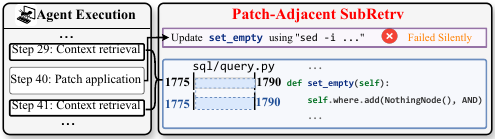}
\caption{Patch-Adjacent SubRetrv of \textit{MSA}\textsubscript{Q35+} on SWE-bench Verified task \code{django-13158}.}
\vspace{-5pt}
\label{fig:app-patch-adj-sr}
\end{wrapfigure}

\textit{Patch-Adjacent SubRetrv}. As illustrated in Figure~\ref{fig:app-patch-adj-sr}, on \code{django-13158}, \textit{MSA\textsubscript{Q35+}} attempts to modify \code{set\_empty} at step 40 using \code{sed -i}, but the edit silently fails. The agent then re-reads the target region at step 41, detects that the function remains unchanged, and subsequently generates a Python script to apply the patch successfully. However, the step-41 retrieval is fully covered by an earlier retrieval at step 29, making it a SubRetrv instance. This example illustrates how limited and error-prone editing support can induce Patch-Adjacent SubRetrv: when an editing tool is difficult to use reliably or provides insufficient feedback about the resulting code state, the agent may re-read previously retrieved code to verify or recover from the edit.

\Needspace*{8\baselineskip}
\begin{wrapfigure}{r}{0.55\textwidth}
\vspace{-12pt}
\centering
\includegraphics[width=\linewidth]{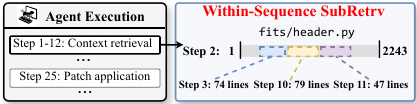}
\caption{Within-Sequence SubRetrv of \textit{MSA}\textsubscript{Q35+} on SWE-bench Verified task \code{astropy-8707}.}
\label{fig:app-within-seq-sr}
\end{wrapfigure}

\textit{Within-Sequence SubRetrv}. As illustrated in Figure~\ref{fig:app-within-seq-sr}, on \code{astropy-8707}, \textit{MSA\textsubscript{Q35+}} reads the entire \code{fits/header.py} (2,243 lines) with \code{cat} at step 2. The agent then zooms back into the same file for specific functions and precise localization during the following retrieval sequence: steps 3, 10, and 11 re-read 74, 79, and 47 lines respectively, each fully contained in the step-2 read.

\Needspace*{10\baselineskip}
\begin{wrapfigure}{r}{0.52\textwidth}
\centering
\includegraphics[width=\linewidth]{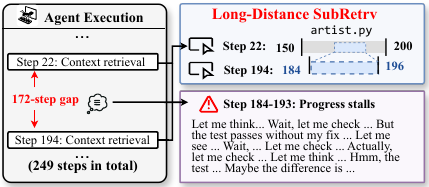}
\caption{Long-Distance SubRetrv of \textit{MSA}\textsubscript{MM3} on SWE-bench Verified task \code{matplotlib-24627}.}
\label{fig:app-long-dist-sr}
\end{wrapfigure}

\textit{Long-Distance SubRetrv}. As illustrated in Figure~\ref{fig:app-long-dist-sr}, on \code{matplotlib-24627}, \textit{MSA\textsubscript{MM3}} reads lines 150--200 of \code{artist.py} at step 22 and re-retrieves lines 184--196 of the same file at step 194, after a 172-step gap. The re-read is triggered by a long debugging loop and a progress stall: at steps 184--193 the agent runs the same pickle test six times while its reasoning repeatedly reverses itself (``Wait, let me check\ldots'', ``Actually, let me check\ldots'', ``Hmm, the test\ldots''). To re-anchor its analysis, the agent re-reads the previously retrieved region, along with regions of \code{test\_pickle.py} and \code{\_version.py} that it had also retrieved before.

\vspace{-6pt}
\subsubsection{Similar Script Generation}
\vspace{-6pt}

\textbf{Classifying script functionalities.}
We classify each SimScrpt script by its function using AST analysis. A script is a \emph{patch script} if it opens a file for writing and transforms its content, through \code{re.sub}, \code{str.replace}, or a read-modify-write sequence. A script is classified as an \emph{assertion test} if it contains \code{assert} statements, imports \code{pytest} or \code{unittest}, or calls \code{self.assert*} methods. A script is an \emph{inspection probe} if it only queries or prints runtime information, such as \code{print}, \code{repr}, and \code{dir} calls, without producing a pass or fail verdict. Scripts matching several signals are assigned by priority: patch, then test, then probe. Scripts that cannot be parsed by AST (0.61--5.54\%) are classified using pattern matching.

\textit{Ephemeral SimScrpt}. The motivating example in Figure~\ref{fig:motivation} already contains this behavior. To reproduce the \code{union().none()} issue on \code{django-13158}, \textit{CC} emits four inline \code{python -c} scripts of 23 to 29 lines. Every script re-emits the same logic: \circled{1} configuring Django settings, \circled{2} initializing the test environment and database, \circled{3} creating test data, \circled{4} building the querysets, and \circled{5} verifying the \code{.none()} behavior that the issue asks to fix. Yet the four scripts differ by only a few lines each: after a missing-table failure in earlier probes, the first script updates the database initialization with \code{migrate --run-syncdb}; the second updates the test-environment setup; the third expands the combined-query test scenarios, replacing \code{count()} with \code{list()} after Django rejects \code{count()} on combined querysets; and the fourth clears the query ordering with \code{order\_by('')} after a default-ordering conflict in the union query. Because each script vanishes after execution, the agent regenerates the identical script logic on every iteration.

\begin{wrapfigure}{r}{0.55\textwidth}
\vspace{-6pt}
\centering
\includegraphics[width=\linewidth]{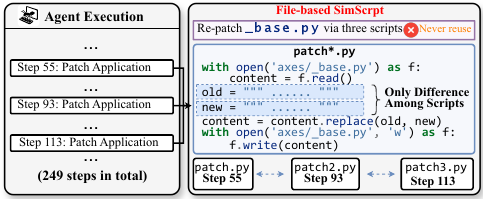}
\caption{File-based SimScrpt of \textit{MSA}\textsubscript{MM3} on SWE-bench Verified task \code{matplotlib-24627}.}
\label{fig:app-file-based-simscrpt}
\end{wrapfigure}

\textit{File-based SimScrpt}. As illustrated in Figure~\ref{fig:app-file-based-simscrpt}, on \code{matplotlib-24627}, \textit{MSA\textsubscript{MM3}} writes \code{/tmp/patch.py} at step 55 to modify \code{axes/\_base.py}, then creates \code{/tmp/patch2.py} at step 93 and \code{/tmp/patch3.py} at step 113 as its fix evolves. The three scripts share the same scaffold, which reads the file, applies a string replacement, and writes the file back; only the strings for replacement change. Although the script already exists on disk, the agent versions its fix by writing a near-duplicate sibling file each time instead of editing the existing one.

Ideally, to prevent the File-Based SimScrpt in Figure~\ref{fig:app-file-based-simscrpt}, the agent would persist a reusable \code{patch.py} that accepts the target file and the old and new strings as parameters. The script should provide explicit feedback on whether the replacement succeeded and report specific errors, such as missing or ambiguous matches. The agent could then reuse the same script for subsequent patches, avoiding repeated script generation and its associated cost.

\subsubsection{Test Re-Execution}

\textbf{Identifying re-executions.}
The executed tests (test ID) are extracted from the commands labeled as \code{Execute_Existing_Tests} and \code{Execute_Generated_Tests} from each inter-patch window. For example, the following commands \code{pytest tests/foo/test\_bar.py}, \code{python -m django test foo.test\_bar}, and \code{python tests/runtests.py foo.test\_bar} map to the same test ID: \code{test\_bar}. Flags and output filters such as \code{-v} or \code{tail} do not affect the extracted tests, and the test IDs of the agent-generated test scripts are recorded by their file path. Within each group with the same test ID, we conservatively retain the final execution as potentially decision-relevant and flag all preceding executions as ReTest.

\begin{wrapfigure}{r}{0.48\textwidth}
\vspace{-12pt}
\centering
\includegraphics[width=\linewidth, trim = {0mm 0mm 0mm 3mm}, clip]{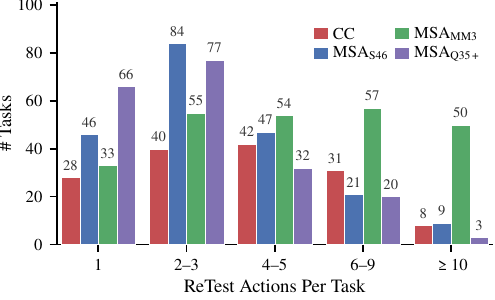}
\captionsetup{skip=2pt}
\caption{Distribution of ReTest actions per task.}
\label{fig:redundant-validation-plot}
\vspace{4pt}
\includegraphics[width=\linewidth]{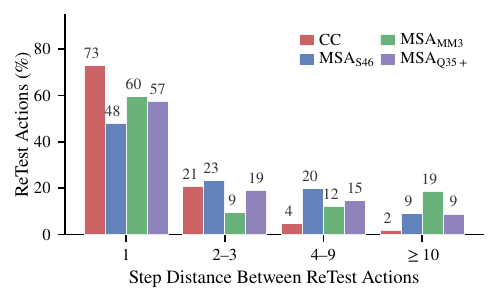}
\caption{Distribution of step distances between ReTest actions within a ReTest group.}
\label{fig:tre-gap}
\vspace{-8pt}
\end{wrapfigure}

Figure~\ref{fig:redundant-validation-plot} shows that ReTest frequency is long-tailed. Depending on the configuration, 23 to 107 affected tasks contain at least six ReTest actions. The tail is most pronounced for \textit{MSA\textsubscript{MM3}}, where 107 of the 249 affected tasks contain at least six ReTest actions and 50 contain at least ten, compared with 39 and 8 of 149 for \textit{CC}. 
Figure~\ref{fig:tre-gap} shows that 48.01--72.84\% of ReTest actions occur at a step distance of one, immediately following the previous execution. \textit{CC} is the most concentrated, with 72.84\% at distance one and only 1.76\% at ten or more, whereas \textit{MSA\textsubscript{MM3}} has the heaviest long-distance tail at 18.78\%. Immediate ReTest is consistent with test knowledge gaps and test-signal recovery, where the agent retries with a different runner or output filter, while long-distance ReTest is consistent with progress stalls. Below, we illustrate the three causes with concrete examples.

\Needspace*{24\baselineskip}

\textit{Repository-specific test knowledge gaps}. Agents may lack knowledge of a repository's expected test runner, working directory, or configuration, leading to repeated failed validations. For example, in Figure~\ref{fig:motivation}, \textit{CC} makes eight failed attempts to run \code{test_qs_combinators}: it repeatedly uses \code{python -m django test}, which cannot resolve the \code{queries} module. \textit{CC} then inspects Django's test structure and discovers its repository-specific test runner. It then invokes \code{tests/runtests.py} with the correct test label, successfully passing all 31 tests.

\begin{wrapfigure}{r}{0.55\textwidth}
\vspace{-12pt}
\centering
\includegraphics[width=\linewidth]{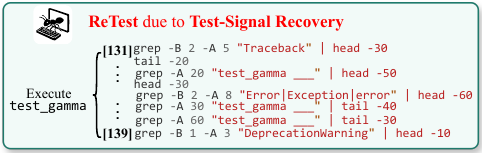}
\caption{Test-signal recovery of \textit{MSA}\textsubscript{MM3} on SWE-bench Verified task \code{sympy-13878}.}
\label{fig:test-signal-recovery}
\end{wrapfigure}

\textit{Test-signal recovery}. A test execution may fail to provide usable signals because its output is truncated, ambiguous, or not effectively captured by the agent, prompting repeated execution to recover the missing information. Such reruns can be locally reasonable because understanding the failure signal is critical for deciding the next debugging step, but they reveal inefficiency in how test results are delivered and retained.
For example, on \code{sympy-13878} (Figure~\ref{fig:test-signal-recovery}), \textit{MSA\textsubscript{MM3}} runs \code{test_gamma} eight times between steps 131--139 while applying different test feedback filters and keyword searches to diagnose two reported exceptions. The failures stem from a Python 3.9 compatibility issue that promotes a \code{DeprecationWarning} to an exception, but \textit{MSA\textsubscript{MM3}} fails to capture an explicit exception marker from the filtered output such as \code{Error}, \code{Traceback}, or \code{Exception}. As a result, the agent repeatedly greps and truncates the test output without isolating the cause. It eventually uses \code{git stash} and reruns the test, successfully identifying the environment-related issue.

\textit{Progress stalls}. In some cases, prior executions already provide clear test signals, yet the agent continues running the same tests without modifying the patch. In these cases, the agent often exhibits decision-making stalls, repeating the same reasoning and validation without advancing the solution. For example, on \code{django-11141}, \textit{MSA\textsubscript{S46}} applies a patch at step 66, with only \code{test_load_empty_dir} failing because it encodes the legacy behavior that the issue intends to remove. The agent recognizes that it should either update this outdated test, which is the correct solution, or redesign its implementation to preserve the old behavior. However, \textit{MSA\textsubscript{S46}} hesitates over whether updating the test functions satisfies the instruction requirements. It consequently reruns the \code{migrations} test suite 22 times and \code{test_makemigrations_no_init} 18 times without another patch until the step budget is exhausted and the task fails.

\vspace{-10pt}
\subsection{Additional Details for RQ2}
\vspace{-6pt}
\label{app:codegraph}

This section provides the additional details for Section~\ref{sec:RQ2}. Appendix~\ref{app:cg-integration} describes why we choose CodeGraph, and how CodeGraph is integrated into the two agent frameworks. Appendix~\ref{app:cg-usage} reports how agents actually use CodeGraph (Table~\ref{tab:app-cg-usage}).

\vspace{-6pt}
\subsubsection{CodeGraph Integration}
\vspace{-6pt}
\label{app:cg-integration}

\textbf{Why CodeGraph.} We choose CodeGraph as a representative tool of structure-aware retrieval for two reasons. First, it is one of the most widely adopted tools of this kind: when we ran our experiments in July 2026, the open-source repository had 63.8K stars on GitHub, and its npm package was downloaded 359K times in that month. Second, it robustly supports integration into both agent frameworks: it ships an MCP server for \textit{CC} and a CLI for \textit{MSA}, and it covers 22 languages in the version we use, including Python, JavaScript, TypeScript, and Go, the languages of the SWE-bench Verified and Pro tasks.

We use CodeGraph v0.9.9~\citep{codegraph}, an open-source code-intelligence tool that parses a repository into an AST-based index of symbols and their relations (definitions, calls, inheritance, and test coverage). The index is built once per task before the agent starts and is refreshed automatically when the agent edits files. CodeGraph is read-only; it never modifies files or runs code. \textit{CC} accesses it through an MCP server that exposes seven functionalities, and \textit{MSA} accesses it through two types of shell commands: \code{cgsrc} (read a symbol's source code by name) and \code{cg} (locate symbols, trace call relationships, estimate change impact, list files). Both harnesses receive equivalent CodeGraph instructions in the system and task prompts. The instructions designate CodeGraph as the primary way to locate and read symbols and to trace call relationships, instead of using \code{grep} or \code{find} for symbol lookup. Ordinary shell commands and tools remain allowed for non-symbol text retrieval such as configuration files, documentation, and error-message strings, and for editing, execution, and other purposes.

\textbf{Task coverage on Pro-100.} CodeGraph cannot run on 13 of the Pro-100 tasks. For 11 of them, the task image is built on Alpine Linux, which uses the musl C library, whereas CodeGraph ships a Node.js binary compiled against glibc, so the binary cannot be executed in these containers. CodeGraph also fails on the remaining two tasks, one from \code{ansible} and one from \code{teleport}. We have not identified the specific cause. We therefore evaluate CodeGraph on the other 87 tasks and compare it with the baseline restricted to the same 87 tasks. On this subset, the baseline Pass@1 is 53.26\%, 49.81\%, 58.62\%, and 42.53\% for \textit{CC}, \textit{MSA\textsubscript{S46}}, \textit{MSA\textsubscript{MM3}}, and \textit{MSA\textsubscript{Q35+}}, respectively.

\newcommand{\appcgcode}[1]{\lstinline[basicstyle=\ttfamily\scriptsize,breaklines=true]@#1@}
\begin{table*}[!htb]
\caption{CodeGraph interfaces of \textit{CC} (MCP tools) and \textit{MSA} (shell commands), using function \code{json\_script} in \code{django/utils/html.py} as an example.}
\label{tab:app-cg-tools}
\centering
\scriptsize
\setlength{\tabcolsep}{4pt}
\renewcommand{\arraystretch}{1.0}
\begin{adjustbox}{max width=\textwidth}
\begin{tabular}{l l l}
\toprule
\textbf{Functionality} & \textbf{\textit{CC} tool call} & \textbf{\textit{MSA} command} \\
\midrule
Read source & \appcgcode{codegraph\_explore(query="json\_script")} & \appcgcode{cgsrc "json\_script"} \\
Locate symbol & \appcgcode{codegraph\_search(query="json\_script")} & \appcgcode{cg query "json\_script"} \\
Read callers & \appcgcode{codegraph\_callers(symbol="json\_script")} & \appcgcode{cg callers "json\_script"} \\
Read callees & \appcgcode{codegraph\_callees(symbol="json\_script")} & \appcgcode{cg callees "json\_script"} \\
Analyze change impact & \appcgcode{codegraph\_impact(symbol="json\_script")} & \makecell[l]{\appcgcode{cg impact "json\_script"}\\\appcgcode{cg affected django/utils/html.py}} \\
Show file structure & \appcgcode{codegraph\_files(path="django/utils")} & \appcgcode{cg files --filter django/utils} \\
Check index status & \appcgcode{codegraph\_status()} & \appcgcode{cg status} \\
\bottomrule
\end{tabular}
\end{adjustbox}
\end{table*}

Table~\ref{tab:app-cg-tools} maps the two interfaces to the underlying functionalities, using the target function \code{json\_script} from the task \code{django-15103} as the example. To read its source code, \textit{CC} calls \code{codegraph\_explore(query="json\_script")} and \textit{MSA} runs \code{cgsrc "json\_script"}; both return the source code of the matching symbols with file paths and line ranges, and \code{codegraph\_explore} additionally appends the callers, relationships, and dependencies of every match. To locate a symbol without retrieving the exact context, \code{codegraph\_search} and \code{cg query} return only its type, location, and signature. For call relationships, \code{codegraph\_callers}/\code{cg callers} and \code{codegraph\_callees}/\code{cg callees} return the names and locations of the function's callers and callees. To analyze what code is affected by changing a symbol, \code{codegraph\_impact} and \code{cg impact} return the symbols that depend on the target symbol, and \code{cg affected} lists the test files that cover the given source file. To show file structure, \code{codegraph\_files} and \code{cg files} return the indexed file tree under a directory together with the programming language and the number of symbols in each file. To check index status, \code{codegraph\_status} and \code{cg status} report the numbers of indexed files, symbols, relations, and whether it is up to date with the working tree.

\subsubsection{CodeGraph Usage}
\label{app:cg-usage}

Table~\ref{tab:app-cg-usage} reports how agents use CodeGraph on Verified-200 and Pro-100. Agents adopt the tool in 74.00--97.50\% of tasks. Reading source accounts for 54.16--98.88\% of CodeGraph queries and locating symbols for another 0.45--36.48\%, which together make up 99.33\% and 89.46\% of \textit{CC}'s queries on Verified-200 and Pro-100 and 80.84--95.30\% of \textit{MSA}'s. Relationship queries (callers, callees, and change impact) receive 0.56--1.60\% of \textit{CC}'s queries and 3.24--7.15\% of \textit{MSA}'s. File-structure queries are rare on Verified-200 (0.11--4.35\%) but more common on Pro-100 (7.32--14.79\%), consistent with the larger, multi-language repositories of SWE-bench Pro.  Index-status queries are almost never issued.

\begin{table*}[t]
\caption{CodeGraph usage in \textbf{RQ2} covering all 100 Pro tasks, including the percentage of tasks in which the agent issued at least one CodeGraph query, and the share of CodeGraph queries by functionality.}
\label{tab:app-cg-usage}
\centering
\scriptsize
\setlength{\tabcolsep}{4pt}
\renewcommand{\arraystretch}{0.8}
\begin{tabular}{l*{8}{c}}
\toprule
& \multicolumn{4}{c}{Verified-200} & \multicolumn{4}{c}{Pro-100} \\
\cmidrule(lr){2-5}\cmidrule(lr){6-9}
& \multicolumn{1}{c}{CC} & \multicolumn{1}{c}{MSA\textsubscript{S46}} & \multicolumn{1}{c}{MSA\textsubscript{MM3}} & \multicolumn{1}{c}{MSA\textsubscript{Q35+}} & \multicolumn{1}{c}{CC} & \multicolumn{1}{c}{MSA\textsubscript{S46}} & \multicolumn{1}{c}{MSA\textsubscript{MM3}} & \multicolumn{1}{c}{MSA\textsubscript{Q35+}} \\
\midrule
CodeGraph task coverage (\%) & 91.50 & 95.00 & 97.50 & 93.17 & 77.00 & 85.00 & 77.00 & 74.00 \\
\midrule
Share of CodeGraph queries (\%) \\
\quad Read source & 98.88 & 73.97 & 69.03 & 58.82 & 81.47 & 63.80 & 64.82 & 54.16 \\
\quad Locate symbol & 0.45 & 20.65 & 18.07 & 36.48 & 7.99 & 17.04 & 19.46 & 31.51 \\
\quad Read callers & 0.56 & 3.62 & 4.28 & 3.30 & 1.60 & 3.24 & 5.36 & 6.29 \\
\quad Read callees & 0.00 & 0.00 & 0.00 & 0.00 & 0.00 & 0.00 & 0.18 & 0.00 \\
\quad Analyze change impact & 0.00 & 0.49 & 2.58 & 0.45 & 0.00 & 0.00 & 1.61 & 0.47 \\
\quad Show file structure & 0.11 & 1.27 & 4.35 & 0.95 & 8.95 & 14.79 & 7.32 & 7.44 \\
\quad Check index status & 0.00 & 0.00 & 1.70 & 0.00 & 0.00 & 1.13 & 1.25 & 0.14 \\
\bottomrule
\end{tabular}
\end{table*}

\subsection{Additional Details for RQ3}
\label{app:rq3}

In this section, we provide additional details on the agent-synthesized and developer-designed skills, including how skills are delivered to the agents, the design of the trajectory analysis agent, the skill synthesis pipeline, and the specific skill contents.

\textbf{Skill delivery.} Agent skills~\citep{anthropic-skills} are usually loaded on demand: the agent initially sees only the skill descriptions and reads the full \texttt{SKILL.md} once it selects a certain skill, which keeps the context small when many skills are available. Our setting involves no skill-selection decision, because each configuration has a single skill set whose rules apply to every coding task. Following Trace2Skill~\citep{Trace2Skill}, we therefore preload the entire skill set into the system prompt. The preloading is complete rather than partial because our skill sets contain only behavioral rules, with no bundled scripts or resources that would otherwise be discovered on demand. Keeping the skill set always present also yields a controlled comparison, since the rules are available on every task rather than conditionally loaded.

\vspace{-5pt}
\subsubsection{Agent-Synthesized Skills}
\label{app:ai-skills}

\textbf{Trajectory analysis agent design.} For the agent-synthesized skills in \textbf{RQ3}, we build a lightweight analysis agent that distills corrective rules from each configuration's RQ1 trajectories using its own backbone model. For each trajectory, the agent receives the task outcome, monetary cost, and step count, together with the RQ1 diagnoses, including the flagged cost-inefficient behavior types, their frequencies, and the flagged step indices. 

\begin{table*}[t]
\centering
\scriptsize
\renewcommand{\arraystretch}{1}
\caption{Toolset of the trajectory-analysis agent.}
\label{tab:analysis-tools}
\begin{tabularx}{0.75\columnwidth}{
    @{}l@{\hspace{1em}}>{\raggedright\arraybackslash}X@{}
}
\toprule
\textbf{Tool} & \textbf{Returns} \\
\midrule
\texttt{get\_skeleton}
& Step indices, action labels, and truncated actions for a trajectory overview. \\

\texttt{get\_step([ids])}
& Full thoughts, commands, and command outputs for selected steps. \\

\texttt{get\_gt\_patch}
& The ground-truth patch of the task. \\

\texttt{get\_agent\_submission}
& The agent's final submitted patch. \\
\bottomrule
\end{tabularx}
\end{table*}
\begin{table*}[t]
\centering
\scriptsize
\renewcommand{\arraystretch}{0.8}
\caption{Cost of synthesizing skills. \textbf{Agent}, \textbf{One-time}, and \textbf{Merge} are the total costs of the agent analysis, the single-call analysis, and the hierarchical consolidation; \textbf{Total}: the total cost of synthesizing skills; \textbf{Task cost}: the cost of solving all 300 RQ1 tasks; \textbf{Ratio}: Total / Task cost.}
\label{tab:skill-synthesis-cost}
\resizebox{0.7\columnwidth}{!}{%
\begin{tabular}{@{}lcccccc@{}}
\toprule
Config & Agent (\$) & One-time (\$) & Merge (\$) & Total (\$) & Task cost (\$) & Ratio \\
\midrule
CC          & 20.33 & 1.78 & 0.65 & 22.76 & 156.00 & 14.59\% \\
MSA\textsubscript{S46}    & 34.48 & 2.03 & 1.08 & 37.59 & 166.06 & 22.64\% \\
MSA\textsubscript{MM3}    &  3.72 & 1.65 & 0.49 &  5.86 & 109.50 &  5.35\% \\
MSA\textsubscript{Q35+} &  2.92 & 1.23 & 0.12 &  4.27 &  30.60 & 13.95\% \\
\bottomrule
\end{tabular}}

\vspace{6pt}
\definecolor{cDown}{HTML}{0072B2}
\definecolor{cUp}{HTML}{D2506A}
\definecolor{cNoise}{HTML}{FFFFFF}
\caption{Effects of three approaches on Pass@1, cost, and CoP. SynSkills is reported with and without synthesis cost (Table~\ref{tab:skill-synthesis-cost}) amortized over Verified-200 and Pro-100 (\textbf{+ synthesis}). {\setlength{\fboxsep}{0pt}\colorbox{cDown!45}{\strut\,blue\,}}/{\setlength{\fboxsep}{0pt}\colorbox{cUp!45}{\strut\,red\,}} indicate robust improvements/regressions, while uncolored cells indicate changes within the run-to-run noise floor; darker shades indicate larger changes.}
\label{tab:app-synskills-amortized}
\centering
\scriptsize
\setlength{\tabcolsep}{2pt}
\renewcommand{\arraystretch}{0.9}
\begin{adjustbox}{max width=1\textwidth}
\begin{tabular}{ll*{8}{r}}
\toprule
& & \multicolumn{4}{c}{Verified-200} & \multicolumn{4}{c}{Pro-100} \\
\cmidrule(lr){3-6}\cmidrule(lr){7-10}
Approach & Metric & \multicolumn{1}{c}{CC} & \multicolumn{1}{c}{MSA\textsubscript{S46}} & \multicolumn{1}{c}{MSA\textsubscript{MM3}} & \multicolumn{1}{c}{MSA\textsubscript{Q35+}} & \multicolumn{1}{c}{CC} & \multicolumn{1}{c}{MSA\textsubscript{S46}} & \multicolumn{1}{c}{MSA\textsubscript{MM3}} & \multicolumn{1}{c}{MSA\textsubscript{Q35+}} \\
\midrule
\multirow{3}{*}{CodeGraph}
 & Pass@1 (pp) & \cellcolor{cNoise}74.17\pct{-0.83} & \cellcolor{cNoise}72.50\pct{+0.00} & \cellcolor{cNoise}75.50\pct{+0.33} & \cellcolor{cDown!18}72.50\pct{+5.33} & \cellcolor{cNoise}55.17\pct{+1.92} & \cellcolor{cNoise}51.72\pct{+1.92} & \cellcolor{cNoise}57.47\pct{-1.15} & \cellcolor{cNoise}39.08\pct{-3.45} \\
 & Cost (\$) & \cellcolor{cNoise}0.596\pct{+8.30\%} & \cellcolor{cNoise}0.619\pct{-3.03\%} & \cellcolor{cUp!36}0.591\pct{+28.14\%} & \cellcolor{cNoise}0.112\pct{-3.29\%} & \cellcolor{cUp!27}1.061\pct{+12.19\%} & \cellcolor{cNoise}1.132\pct{-0.25\%} & \cellcolor{cUp!18}0.985\pct{+8.39\%} & \cellcolor{cUp!27}0.150\pct{+12.99\%} \\
 & CoP & \cellcolor{cNoise}0.804\pct{+10.22\%} & \cellcolor{cNoise}0.854\pct{-3.04\%} & \cellcolor{cUp!36}0.782\pct{+27.52\%} & \cellcolor{cNoise}0.155\pct{-10.63\%} & \cellcolor{cNoise}1.923\pct{+8.69\%} & \cellcolor{cNoise}2.188\pct{-4.32\%} & \cellcolor{cNoise}1.715\pct{+10.20\%} & \cellcolor{cUp!36}0.388\pct{+23.20\%} \\
\midrule
\multirow{3}{*}{SynSkills}
 & Pass@1 (pp) & \cellcolor{cNoise}74.50\pct{-0.50} & \cellcolor{cNoise}72.00\pct{-0.50} & \cellcolor{cNoise}75.50\pct{+0.33} & \cellcolor{cNoise}69.67\pct{+2.50} & \cellcolor{cNoise}55.00\pct{+0.67} & \cellcolor{cNoise}52.00\pct{+0.00} & \cellcolor{cNoise}62.00\pct{+2.00} & \cellcolor{cNoise}46.00\pct{+1.00} \\
 & Cost (\$) & \cellcolor{cNoise}0.541\pct{-1.38\%} & \cellcolor{cDown!36}0.496\pct{-22.32\%} & \cellcolor{cNoise}0.462\pct{+0.15\%} & \cellcolor{cNoise}0.115\pct{-0.70\%} & \cellcolor{cDown!18}0.826\pct{-8.86\%} & \cellcolor{cNoise}1.006\pct{-6.13\%} & \cellcolor{cDown!27}0.766\pct{-11.90\%} & \cellcolor{cNoise}0.132\pct{-2.22\%} \\
 & CoP & \cellcolor{cNoise}0.726\pct{-0.12\%} & \cellcolor{cDown!36}0.689\pct{-21.79\%} & \cellcolor{cNoise}0.611\pct{-0.33\%} & \cellcolor{cNoise}0.166\pct{-4.40\%} & \cellcolor{cDown!18}1.503\pct{-9.48\%} & \cellcolor{cNoise}1.936\pct{-6.40\%} & \cellcolor{cNoise}1.235\pct{-15.09\%} & \cellcolor{cNoise}0.286\pct{-4.51\%} \\
\midrule
\multirow{3}{*}{\makecell[l]{SynSkills\\+ synthesis}}
 & Pass@1 (pp) & \cellcolor{cNoise}74.50\pct{-0.50} & \cellcolor{cNoise}72.00\pct{-0.50} & \cellcolor{cNoise}75.50\pct{+0.33} & \cellcolor{cNoise}69.67\pct{+2.50} & \cellcolor{cNoise}55.00\pct{+0.67} & \cellcolor{cNoise}52.00\pct{+0.00} & \cellcolor{cNoise}62.00\pct{+2.00} & \cellcolor{cNoise}46.00\pct{+1.00} \\
 & Cost (\$) & \cellcolor{cNoise}0.617\pct{+12.51\%} & \cellcolor{cNoise}0.621\pct{-2.70\%} & \cellcolor{cNoise}0.481\pct{+4.39\%} & \cellcolor{cUp!27}0.130\pct{+11.55\%} & \cellcolor{cNoise}0.902\pct{-0.49\%} & \cellcolor{cNoise}1.132\pct{+5.78\%} & \cellcolor{cDown!18}0.785\pct{-9.65\%} & \cellcolor{cUp!18}0.146\pct{+8.48\%} \\
 & CoP & \cellcolor{cNoise}0.828\pct{+13.88\%} & \cellcolor{cNoise}0.863\pct{-2.04\%} & \cellcolor{cNoise}0.637\pct{+3.88\%} & \cellcolor{cNoise}0.186\pct{+7.38\%} & \cellcolor{cNoise}1.641\pct{-1.18\%} & \cellcolor{cNoise}2.176\pct{+5.52\%} & \cellcolor{cNoise}1.266\pct{-12.93\%} & \cellcolor{cNoise}0.317\pct{+6.01\%} \\
\midrule
\multirow{3}{*}{DevSkills}
 & Pass@1 (pp) & \cellcolor{cUp!10}73.50\pct{-1.50} & \cellcolor{cNoise}73.00\pct{+0.50} & \cellcolor{cNoise}73.00\pct{-2.17} & \cellcolor{cNoise}69.83\pct{+2.67} & \cellcolor{cNoise}52.00\pct{-2.33} & \cellcolor{cNoise}52.00\pct{+0.00} & \cellcolor{cNoise}62.00\pct{+2.00} & \cellcolor{cNoise}46.00\pct{+1.00} \\
 & Cost (\$) & \cellcolor{cDown!27}0.473\pct{-13.94\%} & \cellcolor{cDown!46}0.372\pct{-41.73\%} & \cellcolor{cDown!27}0.387\pct{-16.03\%} & \cellcolor{cDown!27}0.103\pct{-11.34\%} & \cellcolor{cNoise}0.894\pct{-1.51\%} & \cellcolor{cDown!46}0.733\pct{-33.00\%} & \cellcolor{cDown!18}0.801\pct{-7.88\%} & \cellcolor{cNoise}0.140\pct{+3.57\%} \\
 & CoP & \cellcolor{cNoise}0.643\pct{-11.83\%} & \cellcolor{cDown!46}0.510\pct{-42.13\%} & \cellcolor{cDown!27}0.530\pct{-13.61\%} & \cellcolor{cDown!27}0.148\pct{-14.83\%} & \cellcolor{cNoise}1.718\pct{+3.58\%} & \cellcolor{cDown!46}1.409\pct{-33.19\%} & \cellcolor{cNoise}1.291\pct{-11.21\%} & \cellcolor{cNoise}0.305\pct{+1.28\%} \\
\bottomrule
\end{tabular}
\end{adjustbox}

\end{table*}

Equipped with four read-only tools (Table~\ref{tab:analysis-tools}), the agent verifies flagged actions against the actual step content in the trajectory and finally returns a list of skill candidates, each pairing a confirmed cost-inefficient anti-pattern with a corrective rule. The agent runs as a ReAct loop~\citep{React} at temperature $0$, with a limit of $8{,}192$ tokens per call and $50$ steps per trajectory.

\textbf{Skill synthesis pipeline.} Running this agent on every trajectory is costly. In preliminary experiments on \textit{MSA\textsubscript{S46}}, the agent tends to read each flagged step and its surrounding context in detail. The number of steps the analysis agent consumes is close to that of the original task-solving trajectory. As a result, the per-trajectory analysis cost is close to the cost of solving the task itself.

To control this cost, we pair the agent with a cheaper single LLM call. We rank each configuration's trajectories by the cost share of their diagnosed cost-inefficient behaviors, analyze the top 25\% with the agent, and process the remaining affected trajectories with a single LLM call, one trajectory at a time. Beyond the inputs given to the agent, the single-call mode additionally receives a compact representation of the trajectory, in which each step is reduced to its action label together with its thought, action, and output truncated to their first and last lines. Finally, we apply the Trace2Skill~\citep{Trace2Skill} hierarchical consolidation to merge the per-trajectory rules into one skill set per configuration.

Table~\ref{tab:skill-synthesis-cost} reports the synthesis cost of each configuration against the cost of solving the corresponding 300 tasks in RQ1. Table~\ref{tab:app-synskills-amortized} amortizes this cost over the 300 evaluation tasks to the per-task cost of SynSkills, and compares the effects of the three mitigating approaches. 
Once amortized, the cost advantage of SynSkills largely disappears: the only robust cost reduction that remains is \textit{MSA\textsubscript{MM3}} on Pro-100 (\mbox{-9.65\%}). \textit{MSA\textsubscript{Q35+}} turns into robust increases on both benchmarks (+11.55\% and +8.48\%), and no setting keeps a robust CoP improvement. Two large savings in Table~\ref{tab:rq234-main-general}, \textit{MSA\textsubscript{S46}} on Verified-200 (\mbox{-22.32\%}) and \textit{CC} on Pro-100 (\mbox{-8.86\%}), shrink to \mbox{-2.70\%} and \mbox{-0.49\%} because these two configurations also have the most expensive synthesis cost (\$37.59 and \$22.76). 
Because DevSkills incur no synthesis cost, they reduce cost more than the amortized SynSkills in seven of the eight settings, the exception being \textit{MSA\textsubscript{MM3}} on Pro-100 (\mbox{-7.88\%} versus \mbox{-9.65\%}). 
However, the skill-synthesis surcharge scales inversely with the number of applied tasks. It would fall below 2.00\% of the baseline cost if applied over 3,000 tasks. So the amortized picture depends on the deployment scale rather than on the skills themselves.

\lstset{
    backgroundcolor=\color{gray!10},
    basicstyle=\ttfamily\scriptsize,
    breaklines=true,
    breakatwhitespace=false,
    columns=fullflexible,
    frame=single,
    numberblanklines=false,
    aboveskip=8pt,
    belowskip=6pt,
    numbers=none,
    literate={→}{{$\rightarrow$}}1 {—}{{\normalfont\textemdash}}1
}

\textbf{Agent-Synthesized Skill Content.}

Claude Code:
\begin{lstlisting}
## Validation & Testing
- Run each test command **once per code state**. Never re-issue an identical command without an intervening code or environment change.
- Before running any test suite, spend one step locating the canonical test runner (check `runtests.py`, `pytest.ini`, `tox.ini`, `Makefile`, `README`, or CI config) and the correct working directory. Use the right invocation on the first try.
- When a command fails, read the full error, diagnose the root cause once, make a concrete corrective change, then retry exactly once. Never re-issue an unchanged failing command.
- After a broader suite passes, skip re-running any subset it already covers.
- Capture full test output in a single run (e.g., verbose flags or `tee`), then apply multiple grep/analysis steps to that captured result rather than re-executing for each filter.
- When a test failure is confirmed as pre-existing, out-of-scope, or environmental, do not re-run overlapping subsets to confirm the same outcome. Move directly to a different approach.
- Once a permission barrier is confirmed (e.g., file owned by root, no sudo), stop probing it. Immediately pivot to known alternative write paths.

## Reading & Navigation
- Read each source file or region **only once per session**. Before issuing any Read or Grep action, check whether the target content was already retrieved and the file is unchanged — reuse cached content.
- Consolidate all required context into a single read, or use targeted grep to find needed line numbers first, then read once. Only re-read when the file has been externally modified or a previously unread section is needed.
- Always read (or confirm you have already read) a file before attempting to edit it. A grep/snippet result alone is insufficient — read the relevant section before issuing an edit.
- When multiple git commands return no meaningful output, stop using git. Switch immediately to reading source files and running reproduction scripts.

## Editing
- Before making any code edit, fully trace all relevant data types, branches, and call sites, then write the complete correct replacement in one edit. Avoid incremental micro-edits that each leave the code broken.
- Before writing any patch, reason through (a) the exact call site that can receive invalid input, (b) the minimal invariant that guards it, and (c) whether fixing the caller vs. callee is safer. Commit to one approach and implement it fully.
- Edit source files in-place using the agent's built-in Edit/Write tool, `sed`, or `patch` rather than creating intermediate copies or numbered variants.
- When using str_replace or patch, ensure the target string exactly matches existing content. After applying any patch, immediately view the modified region to confirm correctness before running tests.
- When a library class or symbol is already confirmed by prior codebase analysis, skip a live import-check command. Trust prior analysis and proceed directly to editing.

## Script & Artifact Hygiene
- Prefer the project's existing test suite over ad-hoc verification scripts. Only create a custom reproduction script if no existing test covers the specific scenario.
- Before writing any script, search the existing codebase first (grep for similar methods, check existing tests, inspect the target function's signature). Existing code usually reveals the correct approach directly.
- Design a **single comprehensive** diagnostic or reproduction script upfront covering all key scenarios (basic case, edge cases, regression). Verify it: (a) imports exist; (b) exercises the code under investigation; (c) output distinguishes "bug present" from "bug fixed". Overwrite in place rather than creating numbered variants.
- Write multi-line verification scripts to a named temporary file (e.g., `/tmp/repro.py`) and execute that file rather than using inline `python -c` commands. Never place it in a directory on `sys.path` or name it after a standard library module. Delete it after use.
- When a script fails for an environmental reason (missing module, wrong path, missing setup), fix the issue in the same script rather than creating a new variant file.
- Use proper assertions (e.g., `assert result == expected`) in verification scripts rather than printing and visually inspecting output.
- Do not run scripts to confirm facts already evident from reading source code, or to re-confirm behavior already validated by a passing test suite.
- Do not write intermediate analysis or notes to temporary files just to read them back immediately. Keep such information in reasoning context.
- When a file-read or directory-listing returns "No such file or directory", verify the directory exists first; if it does not, abandon that path entirely rather than retrying with different filenames.
\end{lstlisting}

Mini-SWE-Agent (Sonnet 4.6):
\begin{lstlisting}
## Repository Navigation & Orientation
- Before invoking any test runner, inspect the repository structure once (README, CI config, Makefile, tox.ini, pytest.ini, runtests.py) to determine the canonical invocation and working directory.
- When a command fails, diagnose the root cause before retrying: distinguish environment/config errors from actual test failures.
- When source code inspection makes behaviour clear, skip writing exploratory scripts to confirm it; read module source directly with `grep`/`sed` instead of launching an interpreter.
- Once a decisive check confirms the required state, stop investigating and report the finding; do not re-examine the same evidence from different angles.

## Editing & Patching
- For small, localized edits use targeted in-place tools (`sed -i` with line numbers, `patch`, `awk`, minimal heredoc, or precise `str_replace`) rather than read-entire-file / write-entire-file cycles.
- Before any substitution, verify the exact target string (e.g., `cat -A` or `print(repr(...))`) to avoid silent no-ops; if two replacement attempts both no-op, switch to a fundamentally different method (line-number-based `sed`/`patch`).
- After applying a patch, verify with a single targeted read (`grep -n`, `sed -n 'START,ENDp'`, or `git diff`) — not a full file re-read.
- Reserve full-file rewrites for changes scattered throughout the file.
- Before making any speculative change, verify with targeted `grep`/`read` that the change is actually needed and that the pattern exists.
- Do not revert a code change unless a concrete test or execution result shows it is wrong; never revert speculatively.
- If a file is accidentally emptied or corrupted, immediately restore it from version control (`git checkout -- <file>`).
- Avoid splitting files into multiple temporary parts to insert a small change; use a single atomic edit to prevent quoting pitfalls and partial-write corruption.

## Script & Artifact Hygiene
- Write one persistent repro/verification script at the start of debugging; consolidate all exploratory, diagnostic, and verification logic into it; edit it in place rather than creating numbered variants.
- Before creating any new script, check whether the desired change is already present (`grep`/`sed`); if so, skip the script entirely.
- For short tasks, prefer `python3 -c` one-liners or heredoc invocations over creating named script files.
- When iterating on a script or patch, overwrite the same file — never create numbered variants.
- Execute a script immediately after writing it; do not overwrite it multiple times without running it.
- When a reproduction script fails to trigger the bug, diagnose the root cause (re-read the issue, inspect the code path) before rewriting — write one comprehensive script covering all cases.
- Use direct shell commands (`cat`, `sed`, `grep`, `git diff`) to inspect content or display diffs rather than writing a temporary Python script.
- When a test fixture or configuration fails, fix the existing fixture in place rather than creating a parallel one; reuse existing helpers.
- For non-ASCII content in shell, use heredoc syntax or a temp script file with `PYTHONIOENCODING=utf-8` rather than inline `-c` commands.

## Validation & Testing
- Run the relevant test suite exactly once per code-change state; only re-run after an intervening code change.
- Capture sufficient diagnostic information in a single well-formed invocation (pipe to `tee`, use `grep -E` for summary lines, or use `-v`) so pass/fail status and details are visible at once.
- Run the fix-verification script once before the fix (to confirm the bug) and once after (to confirm resolution); do not re-run unchanged tests for reassurance.
- Before running a verification script, ensure the fix is complete and handles all known edge cases; enumerate edge cases first, then implement and test once.
- When a large test suite consistently times out, stop retrying with minor flag variations; run targeted subsets relevant to the change instead.
- When verifying pre-existing failures, a single `git stash → run → git stash pop` cycle is sufficient; record the baseline once and compare against it.
- Before running a reproduction script, mentally trace the relevant logic end-to-end and fix all visible errors before the first execution rather than iteratively patching and re-running.
- Before re-reading a file or region, check whether it has been modified since the last read; if unchanged and still in context, skip the re-read entirely.
\end{lstlisting}

Mini-SWE-Agent (MiniMax-M3):
\begin{lstlisting}
## Validation & test execution
- Plan every test, search, or verification command up front: pick verbosity, output format, and filter flags once, then run it once and capture full output (redirect, tee, or a generous window).
- Compose invocation with filtering in a single run (`... | grep ...`, `... | tail -n ...`, or summary flags) rather than full-suite runs followed by re-runs with different post-processing.
- Batch multiple related targets into one invocation; for long runs, redirect full output to a file once and apply all subsequent filters to the saved log.
- Trust passing results: skip post-fix sanity probes (import checks, identity checks, unrelated module reloads) that don't exercise the changed behavior.
- Don't re-invoke the same suite under cosmetic flag changes, different filters, or "just to confirm again." Re-execute only when underlying code or inputs have materially changed.
- Scope regression verification to modules that directly exercise the changed path plus a small set of clearly related ones; defer broader coverage to the project's own CI.

## Scripts & reproduction artifacts
- Maintain one canonical reproduction/verification script per debugging concern; evolve it in place — append cases, parametrize via flags/argv, refine assertions — instead of spawning numbered variants.
- Persist reusable reproduction logic in a single small file on first use, then re-execute that file rather than fighting inline `python -c` invocations with re-imports and re-quoting.
- Consolidate exploratory probes, verification checks, and related cases into one comprehensive script or one batched invocation from the start; reserve inline commands for genuinely disposable single-statement checks.
- If a heredoc-based script keeps failing on quoting, abandon it for a simpler reproduction rather than debugging invisible characters.
- Design repro scripts to be self-contained and leverage existing project infrastructure (in-memory setup, inline model definitions, temporary test methods) instead of fragile standalone scripts that depend on separately-wired app modules or sys.path hacks.
- Clean up temp files at the end of the task.

## File reading & search caching
- Treat content already read or diffed as cached for the rest of the session: never re-cat, re-sed, or re-grep the same unchanged region.
- Rely on `git diff`/`git show` as the authoritative source for post-edit verification; prefer one targeted diff or one narrow read over re-catting the patched region.
- Only re-fetch after a confirmed edit, an external change, truncation, or a genuinely new line range.
- Once a search enumerates and categorizes its results, record the conclusion and move on; if a grep yields nothing relevant, pivot to reading the current code or running the failing test instead of repeating with minor keyword variations.

## Editing approach
- For localized edits (a flag, import, block, or method), apply a surgical in-place patch — narrow anchor-based replace, `sed -i`, or a direct edit tool — rather than reconstructing the whole module via heredoc or full-file overwrite.
- Plan the change by checking surrounding context so the edit lands in one pass without follow-up rewrites; batch multiple coordinated edits into one atomic patch and dry-run against the old content.
- Reserve full-file rewrites for genuinely new files or transformations that genuinely span the file.
- On a small follow-up error, surgically fix the differing line in place instead of restarting the edit.
- When the fix approach is still uncertain, prototype in a scratch buffer before committing to tracked source so speculative edits don't trigger costly round-trips.

## Failure recovery & tooling strategy
- After a small number of failed attempts at the same edit, command, or script, stop iterating and diagnose the root cause (wrong path, missing module, environment issue, escaping problem) via targeted lookup before retrying.
- Statically analyze the code path or read the full error trace to identify all root causes up front and fix them in one revision; avoid the run → see-one-error → fix-one-error → run-again cycle.
- If an edit reports no change or fails identically, commit to a different strategy on the first failure rather than re-issuing the same command.
- Switch tooling (clean heredoc, environment flags like `PYTHONIOENCODING`) rather than re-quoting the same problematic invocation.
- Complete each edit and scratch script in a single direct pass; don't oscillate between write → run → revert → write-new loops or iteratively overwrite a scratch file to fix typos — adjust the script's content itself.

## Exploration discipline
- Before broad searches (filesystem-wide grep/find, git history archaeology, API lookups), verify the hypothesis with cheap local signals: targeted `git blame`, file-scoped `git log`, project-convention check, or a single failing assertion.
- Cap history spelunking at a few focused queries and bound initial exploration to what's needed to pick the edit; defer deep architectural investigation (MRO, class hierarchies, dunder methods, module maps) until after a first attempt if the simpler fix does not suffice.
- Cap broad sweeps (wide `git log --grep`, date-range or keyword variations) to one or two attempts before switching strategy; stop searching once the diff has made the fix clear.
- Once a fix is implemented, verified against a repro, and passing the relevant tests, leave it in place — don't spend steps grepping for ticket numbers or upstream commits, and don't revert a working change to re-derive it from scratch.

## Scope & project conventions
- Define the scope of exploration, edits, and optional artifacts (changelog, docs, supplementary files) before starting; avoid applying-then-reverting speculative changes.
- Cap setup-only operations at one attempt each (single stash/pop baseline, single sleep sized for worst case) and reuse the captured signal for subsequent decisions.
- Before authoring repro or new tests, survey existing project patterns: test layout, bootstrap conventions, runner config, and any prior reproducer for the same code path.
- Read existing tests covering the behavior being changed (including invalid-input/negative cases) and inspect relevant source to confirm attribute names, API shape, and required imports — so iteration focuses on bug behavior rather than `AttributeError`/missing-registration loops.
- When a third-party package seems required, first check whether it is already installed or whether a minimal in-repo stub can exercise the same code path.
- When a task explicitly forbids modifying tests, direct all effort at the source-code fix plus minimal verification that the existing suite still passes; don't invest cycles computing corrected assertion values or applying-then-reverting test edits.

## Output handling
- When output such as a diff or command result must be both inspected and persisted (e.g., a patch deliverable), do it in a single command using `tee` or `> file && cat file`, then reference the saved file at submission.
- Don't write a command's output to a file just to read it back into context — reference the in-context display directly; only persist artifacts actually needed for final submission.

## API & mocking hygiene
- When a public API exposes a callable-based extension parameter (hook, keep_attrs-style), prefer passing the callable over manually reordering positional arguments or post-processing the result — the callable form is the documented extension point and avoids cascading side effects on inferred output metadata.
- Before writing a reproduction that mocks or stubs a class, read the target method once and enumerate every attribute and dunder it references; one targeted read eliminates multi-attempt thrashing caused by missing protocol methods discovered after the fact.

## Version control hygiene
- Use selective version-control operations that target only the files intended to change; avoid broad patterns that may silently discard or overwrite unrelated prior work on other modified files.
\end{lstlisting}

Mini-SWE-Agent (Qwen-3.5 Plus):
\begin{lstlisting}
## Context & Navigation
- Cache retrieved file content, search results, and verified patterns; reference this context instead of re-reading unless the file has been modified or a specific missing section is required.
- Navigate directly to target lines or use comprehensive search queries immediately, avoiding preliminary reads of unrelated sections or fragmented searches.
- Consolidate verification into a single read operation after applying multiple sequential edits to the same region, rather than reading after each individual change.
- Use targeted search commands with specific patterns instead of broadly listing files or investigating features not strictly required by the task.

## Editing & Artifact Hygiene
- Modify existing files in-place using targeted, surgical edits (e.g., line-specific replacements) instead of rewriting entire files or recreating them from scratch.
- Consolidate all related test scenarios, investigations, and fixes into a single file, avoiding the creation of multiple numbered variants, separate scripts, or sequential patch artifacts.
- Apply code modifications directly using in-place editing tools instead of creating temporary wrapper scripts, intermediate patch files, or numbered variants.
- Prefer direct programmatic editing over chained shell commands for complex changes; if a command fails, analyze the error or fall back to manual logic immediately rather than retrying broken syntax.
- Plan complete directory layouts and specify explicit text encoding when creating scripts to generate all required files in a single pass.
- Verify files are tracked by version control and relevant to the deliverable before modifying, skipping build artifacts, cache directories, or untracked files.
- Avoid unnecessary git operations like stash/pop cycles when the working state is correct, and review the final diff exactly once after all changes are complete.

## Validation & Testing
- Execute validation commands only once per code state using minimal, focused test cases that directly validate specific hypotheses, capturing full output to avoid re-execution.
- Consolidate multiple test scenarios, edge cases, verification variants, and diagnostic details into a single comprehensive command or script execution.
- When validation fails, diagnose the root cause and extract all necessary information from that single run to fix errors immediately, rather than attempting multiple similar invocations.
- Establish a baseline for pre-existing failures and verify environment dependencies early by running the full test suite once before applying fixes.
- Skip re-running validation, reverting changes, or repeating tests if the underlying code has not changed, the state is confirmed via inspection, or static verification suffices.
- Once an issue is reproduced or the root cause identified, proceed directly to fix implementation without additional reproduction attempts or consecutive verification steps.
- Optimize search and verification by combining related patterns into single operations, scoping checks strictly to modified files, and trusting targeted edits succeeded unless an error is reported.
- Specify explicit encodings or use automatic handling tools from the first attempt when processing source code to prevent encoding-related retries.

## Diagnostic Efficiency
- Prioritize static inspection of source code, diffs, and environment factors to answer questions before executing ad-hoc diagnostic commands or creating debug artifacts.
- Consolidate multiple exploratory checks and diagnostic commands into single comprehensive scripts or batched invocations, removing temporary artifacts immediately after use.
- Explicitly document key findings from failed iterations to prevent repeating unproductive patterns.
- Verify the existence and accessibility of all required output artifacts and ensure the test environment is fully initialized before initiating final submission.
\end{lstlisting}

\subsubsection{Full Text of the Developer-Designed Skills}
\label{app:human-skills}

The developer-designed skills are listed below.

\lstset{
    backgroundcolor=\color{gray!10},
    basicstyle=\ttfamily\scriptsize,
    breaklines=true,
    breakatwhitespace=false,
    columns=fullflexible,
    frame=single,
    numberblanklines=false,
    aboveskip=8pt,
    belowskip=6pt,
    numbers=none
}

{
\begin{lstlisting}
1. Hypothesis-Before-Read
Before every file read, state in one concise sentence: What you expect to find and why. If you cannot state a specific expectation, reason from existing context first. Do not read blindly.

2. Reuse Context
Before reading content (`grep / cat / sed / ...`), check whether it is already available in the current context. If yes, reuse it instead of fetching it again. Re-read only when necessary, and briefly state why (e.g., "confirming exact string before edit", "content is too far back in context to rely on"). Avoid issuing multiple overlapping reads to retrieve the same context via different tools.

3. Targeted Patch Only
When modifying source code:
- **DO NOT use**: `cat > file << 'EOF'` or `open(f,'w').write("""...""")` to overwrite an entire existing file.
- **Use**: Targeted editing, such as `sed -i` or creating your own python script (`content.replace(old_substr, new_substr)`, triple-quoted for multi-line).
Write the minimal change. Identify the exact old string and new string before patching.

4. Understand the Test Harness Before Blindly Running Tests
Identify the repository test mechanism before running tests / after running tests with wrong commands:
- Check existing documentation / CI configs / test scripts for the correct runner, environment variables, and test syntax.
- Treat import, configuration, dependency, and test-loader failures as harness errors, not patch failures.
Skip this if you already know the working command.

5. Capture Test Output; Filter, Don't Rerun
When running a test, redirect full output to a file, such as:
```
<test_command> > test_out.txt 2>&1
```
Extract test signal precisely from a prior run instead of re-running the test. Only rerun if the source code has changed since the last run.

6. Persist Artifacts; Version Only With Intent
When creating reproduction, test, or patch scripts, save reusable or multi-step logic as a workspace file; use inline commands only for simple, one-off probes. Do not create numbered variants such as `repro2.py` or `fix_v3.py`; revise the original artifact unless separate versions are necessary. State the reason before creating.

7. Break Loops
When repeated reads, tests, or edits produce no new evidence, **Stop.** You may be stuck. Do not repeat the same action and state what remains uncertain, form a concrete hypothesis, then take one materially different action to move on.

If you are stuck on sub-problems (e.g., verification issues like wrong test environment setup), submit your best current solution. A correct patch you cannot fully verify is better than a budget exhausted without submission.
\end{lstlisting}
}

\end{document}